\documentclass{article}
\usepackage[utf8]{inputenc}
\pdfoutput=1
\usepackage[sc,osf]{mathpazo}
\usepackage[height=8.5in,width=6in,letterpaper]{geometry}
\usepackage[sort&compress,numbers]{natbib}
\usepackage[colorlinks=true,citecolor=blue,breaklinks]{hyperref}
\usepackage[hyphenbreaks]{breakurl} 
\usepackage[tracking]{microtype}

\makeatletter
\newlength\aftertitskip     \newlength\beforetitskip
\newlength\interauthorskip  \newlength\aftermaketitskip

\def\maketitle{\par
 \begingroup
   \def\thefootnote{\fnsymbol{footnote}}
   \def\@makefnmark{\hbox to 4pt{$^{\@thefnmark}$\hss}}
   \@maketitle \@thanks
 \endgroup
\setcounter{footnote}{0}
 \let\maketitle\relax \let\@maketitle\relax
 \gdef\@thanks{}\gdef\@author{}\gdef\@title{}\let\thanks\relax}

\def\@startauthor{\noindent \normalsize\bf}
\def\@endauthor{}
\def\@starteditor{\noindent \small {\bf Editor:~}}
\def\@endeditor{\normalsize}
\def\@maketitle{\vbox{\hsize\textwidth
 \linewidth\hsize \vskip \beforetitskip
 {\begin{center} \LARGE\@title \par \end{center}} \vskip \aftertitskip
 {\def\and{\unskip\enspace{\rm and}\enspace}%
  \def\addr{\small\it}%
  \def\email{\hfill\small\tt}%
  \def\name{\normalsize\bf}%
  \def\AND{\@endauthor\rm\hss \vskip \interauthorskip \@startauthor}
  \@startauthor \@author \@endauthor}
}}

\makeatother

\pdfoutput=1                    

\usepackage{floatrow} 
\newfloatcommand{capbtabbox}{table}[][\FBwidth]

\usepackage{graphicx}
\usepackage{amsmath,amsthm,amssymb,bm} 
\usepackage{amsfonts}
\usepackage{mathrsfs}
\usepackage{caption}
\usepackage{subcaption}
\usepackage{multirow}
\usepackage{xspace}
\usepackage{array}
\usepackage{enumerate}
\usepackage{algorithm}
\usepackage{algorithmicx,algpseudocode}
\usepackage{hyperref}
\usepackage{stmaryrd}
\usepackage{appendix}
\usepackage{wrapfig}
\numberwithin{equation}{section}

\newcommand{\reals}{\mathbf{R}}

  \newcommand{\cc}{\mathcal{C}}     \newcommand{\ch}{\mathcal{H}} \newcommand{\ci}{\mathcal{I}}    \newcommand{\cm}{\mathcal{M}}  \newcommand{\co}{\mathcal{O}} \newcommand{\cp}{\mathcal{P}}         

\newcommand{\argmax}{\textrm{argmax}}

\theoremstyle{plain}
\newtheorem{theorem}{Theorem}
\newtheorem{corollary}[theorem]{Corollary}

\newtheorem{lemma}[theorem]{Lemma}

\theoremstyle{definition}
\newtheorem{definition}[theorem]{Definition}

\theoremstyle{remark}

\newcommand{\eps}{\varepsilon}

\newcommand{\dpp}{\textsc{Dpp}\xspace}

\newcommand{\prb}{\pi_{\cc}}

\title{Fast Mixing Markov Chains for Strongly Rayleigh Measures, DPPs, and Constrained Sampling}
\author{\name Chengtao Li \email{ctli@mit.edu}\\
  \name Stefanie Jegelka \email{stefje@csail.mit.edu}\\
  \name Suvrit Sra \email{suvrit@mit.edu}\\
  \addr{Massachusetts Institute of Technology}
}

\begin{document}

\maketitle

\begin{abstract}
  We study probability measures induced by set functions with constraints. Such measures arise in a variety of real-world settings, where prior knowledge, resource limitations, or other pragmatic considerations impose constraints. We consider the task of rapidly sampling from such constrained measures, and develop fast Markov chain samplers for them. Our first main result is for MCMC sampling from Strongly Rayleigh (SR) measures, for which we present sharp polynomial bounds on the mixing time. As a corollary, this result yields a fast mixing sampler for Determinantal Point Processes (DPPs), yielding (to our knowledge) the first provably fast MCMC sampler for DPPs since their inception over four decades ago. Beyond SR measures, we develop MCMC samplers for probabilistic models with hard constraints and identify sufficient conditions under which their chains mix rapidly. We illustrate our claims by empirically verifying the dependence of mixing times on the key factors governing our theoretical bounds. 
\end{abstract}

\section{Introduction}
\vspace*{-8pt}
Distributions over subsets of objects arise in a variety of machine learning applications. They  occur as discrete probabilistic models \cite{bouchard10,smith08,zhang2015higher,gps89,kulesza2012determinantal} in computer vision, computational biology and natural language processing. They also occur in combinatorial bandit learning \cite{cesabianchi09}, as well as in recent applications to neural network compression \cite{mariet16} and matrix approximations \cite{li2016fast}. 

Yet, practical use of discrete distributions can be hampered by  computational challenges due to their combinatorial nature. Consider for instance sampling, a task fundamental to learning, optimization, and approximation. Without further restrictions, efficient sampling can be impossible \cite{dyer99}. 
Several lines of work thus focus on identifying tractable sub-classes, which in turn have had wide-ranging impacts on modeling and algorithms. 
Important examples include the Ising model \cite{jerrum93ising}, matchings (and the matrix permanent) \cite{jerrum04}, spanning trees (and graph algorithms) \cite{broder1989generating,frieze14,spielman08,anari15}, and Determinantal Point Processes (\dpp{}s) that have gained substantial attention in machine learning~\citep{kulesza2012determinantal,li2016gauss,gartrell16,kang2013fast,anari2016monte,kojima2014determinantal}.


In this work, we extend the classes of tractable discrete distributions.  Specifically, we consider the following two classes of distributions on $2^V$ (the set of subsets of a ground set $V = [N] := \{1,\ldots, N\}$): (1) strongly Rayleigh (SR) measures, and (2) distributions with certain cardinality or matroid-constraints. We analyze Markov chains for sampling from both classes. As a byproduct of our analysis, we answer a long-standing question about rapid mixing of MCMC sampling from DPPs. 

SR measures are defined by strong negative correlations, and have recently emerged as valuable tools in the design of algorithms~\citep{anari15}, in the theory of polynomials and combinatorics~\citep{borcea2009negative}, and in machine learning through \dpp{}s, a special case of SR distributions. Our first main result is the first polynomial-time sampling algorithm that applies to all SR measures (and thus \emph{a fortiori} to \dpp{}s).

General distributions on $2^V$ with constrained support (case (2) above) typically arise upon incorporating prior knowledge or resource constraints. We focus on resource constraints such as bounds on cardinality and bounds on including limited items from sub-groups. 
Such constraints can be phrased as a family $\cc \subseteq 2^V$ of subsets; we say $S$ satisfies the constraint $\cc$ iff $S\in\cc$. Then the distribution of interest is of the form
\begin{equation}\label{eq:dist}
  \pi_{\cc}(S) \propto \exp(\beta F(S)) \llbracket S \in \cc\rrbracket,
\end{equation}
where $F: 2^V\to \reals$ is a set function that encodes relationships between items $i \in V$, $\llbracket \cdot \rrbracket$ is the Iverson bracket, and $\beta$ a constant (also referred to as the inverse \emph{temperature}). Most prior work on sampling with combinatorial constraints (such as sampling the bases of a matroid), assumes that $F$ breaks up linearly using element-wise weights $w_i$, i.e., $F(S) = \sum_{i \in S}w_i$. In contrast, we allow generic, nonlinear functions, and obtain a mixing times governed by structural properties of $F$.

\paragraph{Contributions.} We briefly summarize the key contributions of this paper below. 
\vspace{-5pt}
\begin{list}{–}{\leftmargin=1em}\setlength{\itemsep}{0pt}
\item  
We derive a provably fast mixing Markov chain for efficient sampling from strongly Rayleigh measure $\pi$ (Theorem~\ref{thm:rayleigh}). 
This Markov chain is novel and may be of independent interest. 
Our results provide the first polynomial guarantee (to our knoweldge) for Markov chain sampling from a general \dpp, and more generally from an SR distribution.\footnote{The analysis in \cite{kang2013fast} is not correct since it relies on a wrong construction of path coupling.}
\item  We analyze (Theorem~\ref{thm:matroid}) mixing times of an exchange chain 
  when the constraint family $\cc$ is the set of bases of a special matroid, i.e., $|S|=k$ or $S$ obeys a partition constraint. Both of these constraints have high practical relevance~\cite{kulesza2011k,kathuria2016sampling,zhang2015higher}. 
\item We analyze (Theorem~\ref{thm:size}) mixing times of an add-delete chain 
  for the case $|S| \leq k$, which, perhaps surprisingly, turns out to be quite different from $|S|=k$. This constraint can be more practical than the strict choice $|S|=k$, because in many applications, the user may have an upper bound on the budget, but may not necessarily want to expend all $k$ units.
\end{list}
\vspace{-5pt}
Finally, a detailed set of experiments illustrates our theoretical results.

\paragraph{Related work.} 
Recent work in machine learning addresses sampling from distributions with sub- or supermodular $F$~\cite{gotovos2015sampling,rebeschini2015fast}, determinantal point processes \cite{anari2016monte,li2016fast}, and sampling by optimization~\cite{ermon2013embed,a2013maddison}. 
Many of these works (necessarily) make additional assumptions on $\pi_{\cc}$, or are approximate, or cannot handle constraints. Moreover, the constraints cannot easily be included in $F$: an out-of-the-box application of the result in~\cite{gotovos2015sampling}, for instance, would lead to an unbounded constant in the mixing time.


Apart from sampling, other related tracts include work on variational inference for combinatorial distributions \cite{bouchard10,djolonga2014map,smith08,zhang2015higher} and inference for submodular processes \cite{iyer2015submodular}. 
Special instances of~\eqref{eq:dist} include~\cite{kulesza2011k}, where the authors limit \dpp{}s to sets that satisfy $|S|=k$; partition matroid constraints are studied in~\cite{kathuria2016sampling}, while the budget constraint $|S|\le k$ has been used recently in learning \dpp{}s~\citep{gartrell16}. 
Important existing results show fast mixing for a sub-family of strongly Rayleigh distributions \cite{feder1992balanced,anari2016monte}; but those results do not include, for instance, general \dpp{}s.

\subsection{Background and Formal Setup}
Before describing the details of our new contributions, let us briefly recall some useful background that also serves to set the notation. Our focus is on sampling from $\pi_{\cc}$ in~\eqref{eq:dist}; 
we denote by $Z=\sum_{S\subseteq V}\exp(\beta F(S))$ and $Z_\cc=\sum_{S\subseteq \cc}\exp(\beta F(S))$.
The simplest example of $\pi_{\cc}$ is the uniform distribution over sets in $\cc$, where $F(S)$ is constant. In general, $F$ may be highly nonlinear.

We sample from $\pi_{\cc}$ using MCMC, i.e., we run a Markov Chain with state space $\cc$. All our chains are ergodic. The \emph{mixing time} of the chain indicates the number of iterations $t$ that we must perform (after starting from an arbitrary set $X_0 \in \cc$) before we can consider $X_t$ as a valid sample from $\pi_{\cc}$. Formally, if $\delta_{X_0}(t)$ is the total variation distance between the distribution of $X_t$ and $\pi_{\cc}$ after $t$ steps, then $\tau_{X_0}(\eps) = \min\{t: \delta_{X_0}(t')\le \eps,\ \forall t'\ge t\}$ is the mixing time to sample from a distribution $\epsilon$-close to $\pi_{\cc}$ in terms of total variation distance. We say that the chain mixes fast if $\tau_{X_0}$ is polynomial in $N$.
The mixing time can be bounded in terms of the eigenvalues of the transition matrix, as the following classic result shows:
\begin{theorem}[Mixing Time~\cite{diaconis1991geometric}] 
Let $\lambda_i$ be the eigenvalues of the transition matrix, and $\lambda_{\max} = \max\{\lambda_2, |\lambda_N|\} < 1$. Then, the mixing time 
starting from an initial set $X_0\in \cc$ is bounded as
\begin{align*}
\tau_{X_0}(\eps)\le (1 - \lambda_{\max})^{-1}(\log\pi_\cc(X_0)^{-1} + \log\eps^{-1}).
\end{align*}
\end{theorem}
Most of the effort in bounding mixing times hence is devoted to bounding this eigenvalue.

\section{Sampling from Strongly Rayleigh Distributions}
\label{sec:rayleigh}
In this section, we consider sampling from \emph{strongly Rayleigh (SR)} distributions. Such distributions capture the strongest form of negative dependence properties, while enjoying a host of other remarkable properties~\cite{borcea2009negative}. For instance, they include the widely used \dpp{}s as a special case. A distribution is SR if its generating polynomial $p_{\pi}: \mathbb{C}^N \to \mathbb{C}$, $p_{\pi}(z) = \sum_{S \subseteq V} \pi(S) \prod_{i \in S}z_i$ is \emph{real stable}. This means if $\Im(z_i)>0$ for all arguments $z_i$ of $p_{\pi}(z)$, then $p_{\pi}(z)>0$.

We show in particular that SR distributions are amenable to efficient Markov chain sampling. Our starting point is the observation of~\cite{borcea2009negative} on  closure properties of SR measures; of these we use \emph{symmetric homogenization}. Given a distribution $\pi$ on $2^{[N]}$, its symmetric homogenization $\pi_{sh}$ on $2^{[2N]}$ is
\begin{align*}
\pi_{sh}(S) := \left\{\begin{array}{cc}
\pi(S\cap [N]){N \choose S\cap [N]}^{-1} & \text{if } |S| = N;\\
0 & \text{otherwise}.
\end{array}
\right.
\end{align*}
If $\pi$ is SR, so is $\pi_{sh}$. We use this property below in our derivation of a fast-mixing chain.

We use here a recent result of~\citet{anari2016monte}, who show a Markov chain   that mixes rapidly for \emph{homogeneous SR} distributions. These distributions are over all subsets $S \subseteq V$ of some fixed size $|S|=k$, and hence do not include general \dpp{}s. Concretely, for any $k$-homogeneous SR distribution $\pi: \{0,1\}^{N}\to \mathbb{R}_+$, a Gibbs-exchange sampler has mixing time
\begin{align*}
\tau_{X_0}(\eps)\le 2k(N-k)(\log\pi(X_0)^{-1} + \log\eps^{-1}).
\end{align*}
This sampler uniformly samples one item in the current set, and one outside the current set, and swaps them with an appropriate probability.
Using these ideas we show how to obtain fast mixing chains for \emph{any} general SR distribution $\pi$ on $[N]$. First, we construct its symmetric homogenization $\pi_{sh}$, and sample from $\pi_{sh}$ using a Gibbs-exchange sampler. This chain is fast mixing, thus we will efficiently get a sample $T\sim \pi_{sh}$. The corresponding sample for $\pi$ can be then obtained by computing $S = T\cap V$. Theorem~\ref{thm:rayleigh}, proved in the appendix, formally establishes the validity of this idea.
\begin{theorem}\label{thm:rayleigh}
If $\pi$ is SR, then the mixing time of a Gibbs-exchange sampler for $\pi_{sh}$ 
is bounded as
\begin{align}\label{eq:rayleighmix}
\tau_{X_0}(\eps)\le 2 N^2 \Bigl(\log {N\choose |X_0|} + \log (\pi(X_0))^{-1} + \log \eps^{-1}\Bigr).
\end{align}
\end{theorem}
For Theorem~\ref{thm:rayleigh} we may choose the initial set 
such that $X_0$ makes the first term in the sum logarithmic in $N$ ($X_0 = T_0 \cap V$ in Algorithm \ref{algo:rayleigh}).

\begin{algorithm}\small
	\caption{\small Markov Chain for Strongly Rayleigh Distributions}\label{algo:rayleigh}
	\begin{algorithmic} 
	\Require{SR distribution $\pi$}
	\State Initialize $T\subseteq[2N]$ where $|T| = N$ and take $S = T\cap V$
	\While{not mixed}
		\State Draw $q\sim \text{Unif }[0,1]$
		\State Draw $t\in V\backslash S$ and $s\in S$ uniformly at random
		\If{$q\in[0,{(N-|S|)^2\over 2N^2})$} 
			\State $S = S\cup\{t\}$ with probability $\min\{1,{\pi(S\cup\{t\})\over \pi(S)}\times {|S|+1\over N-|S|}\}$ \Comment{Add $t$}
		\ElsIf{$q\in [{(N-|S|)^2\over 2N^2},{N-|S|\over 2N})$}
			\State $S = S\cup\{t\}\backslash\{s\}$ with probability $\min\{1,{\pi(S\cup\{t\}\backslash\{s\})\over \pi(S)}\}$ \Comment{Exchange $s$ with $t$}
		\ElsIf{$q\in [{N-|S|\over 2N}, {|S|^2 + N(N-|S|)\over 2N^2})$}
			\State $S = S\backslash\{s\}$ with probability $\min\{1,{\pi(S\backslash\{s\})\over \pi(S)}\times {|S|\over N-|S|+1}\}$ \Comment{Delete $s$}
		\Else
			\State Do nothing
		\EndIf
	\EndWhile
\end{algorithmic}
\end{algorithm}

\textbf{Efficient Implementation.} Directly running a chain to sample $N$ items from a (doubled) set of size $2N$ 
adds some computational overhead. Hence, we construct an equivalent, more space-efficient chain (Algorithm~\ref{algo:rayleigh}) on the initial ground set $V=[N]$ that only manintains $S\subseteq V$. Interestingly, this sampler is a mixture of add-delete and Gibbs-exchange samplers. 
%
This combination makes sense intuitively, too: add-delete moves~(also shown in Alg.~\ref{algo:scmc}) are needed since the exchange sampler cannot change the cardinality of $S$. But a pure add-delete chain can stall if the sets concentrate around a fixed cardinality (low probability of a larger or smaller set). Exchange moves will not suffer the same high rejection rates.
The key idea underlying Algorithm~\ref{algo:rayleigh} is that the elements in $\{N+1,\ldots, 2N\}$ are indistinguishable, so it suffices to maintain merely the cardinality of the currently selected subset instead of all its indices. Appendix~\ref{app:rayleigh} contains a detailed proof.

\begin{corollary} \label{cor:rayleigh}
  The bound~\eqref{eq:rayleighmix} applies to the mixing time of Algorithm~\ref{algo:rayleigh}. 
\end{corollary}

\textbf{Remarks.} By assuming $\pi$ is SR, we obtain a clean bound for fast mixing. Compared to the bound in~\cite{gotovos2015sampling}, our result avoids the somewhat opaque factor $\exp(\beta\zeta_F)$ that depends on $F$. 

In certain cases, the above chain may mix slower in practice than a pure add-delete chain that was used in previous works \cite{kang2013fast,gotovos2015sampling}, since its probability of doing nothing is higher. In other cases, it mixes much faster than the pure add-delete chain; we observe both phenomena in our experiments in Sec.~\ref{sec:experiments}. Contrary to a simple add-delete chain, in all cases, it is \emph{guaranteed} to mix well. 


\vspace*{-5pt}
\section{Sampling from Matroid-Constrained Distributions}
\label{sec:matroids}
\vspace*{-5pt}
In this section we consider sampling from an explicitly-constrained distribution  $\pi_\cc$ where $\cc$ specifies certain matroid base constraints (\S\ref{sec:mbase}) or a uniform matroid of a given rank (\S\ref{sec:umat}).

\subsection{Matroid Base Constraints}
\label{sec:mbase}
\vspace*{-5pt}
We begin with constraints that are special cases of matroid bases\footnote{Drawing even a uniform sample from the bases of an arbitrary matroid can be hard.}: 
\begin{enumerate}\setlength{\itemsep}{-2pt}
\item \emph{Uniform matroid: } $\cc = \{S \subseteq V \mid |S|=k\}$,
\item \emph{Partition matroid: } Given a partition $V = \bigcup_{i=1}^k \cp_i$, we allow sets that contain exactly one element from each $\cp_i$: $\cc = \{S \subseteq V \mid |S \cap \cp_i| = 1 \text{ for all } 1 \leq i \leq k\}$.
\end{enumerate}
An important special case of a distribution with a uniform matroid constraint is the  $k$-DPP~\cite{kulesza2011k}. Partition matroids are used in multilabel problems \cite{zhang2015higher}, and also in probabilistic diversity models~\cite{iyer2015submodular}.




\begin{algorithm}[h!]
\begin{small}
	\caption{\small Gibbs Exchange Sampler for Matroid Bases}\label{algo:mbmc}
	\begin{algorithmic} 
	\Require{set function $F$, $\beta$, matroid $\cc\subseteq 2^V$}
	\State Initialize $S\in\cc$
	\While{not mixed}
		\State Let $b = 1$ with probability $0.5$
		\If{$b = 1$} 
			\State Draw $s\in S$ and $t\in V\backslash S$ ($t \in \cp(s)\setminus\{s\}$) uniformly at random
			\If{$S\cup\{t\}\backslash\{s\}\in\cc$}
				\State $S\leftarrow S\cup\{t\}\backslash\{s\}$ with probability ${\pi_\cc(S\cup\{t\}\backslash\{s\})\over \pi_\cc(S) + \pi_\cc(S\cup\{t\}\backslash\{s\})}$
			\EndIf
		\EndIf
	\EndWhile
\end{algorithmic}
\end{small}
\end{algorithm}

The sampler is shown in Algorithm~\ref{algo:mbmc}. At each iteration, we randomly select an item $s\in S$ and $t\in V\backslash S$ such that the new set $S\cup\{t\}\backslash\{s\}$ satisfies $\cc$, and swap them with certain probability. For uniform matroids, this means $t\in V\backslash S$; for partition matroids, $t \in \cp(s)\setminus\{s\}$ where $\cp(s)$ is the part that $s$ resides in. The fact that the chain has stationary distribution $\pi_\cc$ can be inferred via detailed balance. Similar to the analysis in \cite{gotovos2015sampling} for \emph{unconstrained} sampling, the mixing time depends on a quantity that measures how much $F$ deviates from linearity: $\zeta_F = \max_{S,T\in\cc}|F(S) + F(T) - F(S\cap T) - F(S\cup T)|$. Our proof, however, differs from that of \cite{gotovos2015sampling}. While they use canonical paths~\cite{diaconis1991geometric}, we use multicommodity flows, which are more effective in our constrained setting.


\begin{theorem}\label{thm:matroid}
Consider the chain in Algorithm~\ref{algo:mbmc}. For the uniform matroid, $\tau_{X_0}(\eps)$ is bounded as
\begin{align}\label{eq:uniformbase}
  \tau_{X_0} (\eps)\le 4k(N-k)\exp(\beta(2\zeta_F))(\log\pi_\cc(X_0)^{-1} + \log\eps^{-1});
\end{align}
For the partition matroid, the mixing time is bounded as
\begin{align}
\tau_{X_0} (\eps)\le 4k^2\max_i |\cp_i|\exp(\beta(2\zeta_F))(\log\pi_\cc(X_0)^{-1} + \log\eps^{-1}).
\end{align}
%
%
\end{theorem}
Observe that if $\cp_i$'s form an equipartition, i.e., $|\cp_i| = N/k$ for all $i$, then the second bound becomes $\widetilde{\co}(kN)$. For $k = \co(\log N)$, the mixing times depend as $\co(N \mathrm{polylog}(N)) = \widetilde{\co}(N)$ on $N$. For uniform matroids, the time is equally small if $k$ is close to $N$. Finally, the time depends on the initialization, $\prb(X_0)$. If $F$ is monotone increasing, one may run a simple greedy algorithm to ensure that  $\prb(X_0)$ is large. If $F$ is monotone submodular, this ensures that $\log \prb(X_0)^{-1} = \co(\log N)$.

Our proof uses a multicommodity flow to upper bound the largest eigenvalue of the transition matrix. Concretely, let $\ch$ be the set of all simple paths between states in the state graph of Markov chain, we construct a flow $f:\ch\to\mathbb{R}^+$ that assigns a nonnegative flow value to any simple path between any two states (sets) $X, Y \in \cc$. Each edge $e = (S,T)$ in the graph has a capacity $Q(e) = \prb(S)P(S,T)$ where $P(S,T)$ is the transition probability from $S$ to $T$. The total flow sent from $X$ to $Y$ must be $\prb(X)\prb(Y)$: if $\ch_{XY}$ is the set of all simple paths from $X$ to $Y$, then we need $\sum_{p \in \ch_{XY}} f(p) = \prb(X)\prb(Y)$. Intuitively, the mixing time relates to the congestion in any edge, and the length of the paths. If there are many short paths $X \rightsquigarrow Y$ across which flow can be distributed, then mixing is fast. This intuition is captured in a fundamental theorem:

\begin{theorem}[Multicommodity Flow~\cite{sinclair1992improved}]\label{thm:multiflow}
  Let $E$ be the set of edges in the transition graph, and $P(X,Y)$ the transition probability. Define
 \begin{align*}\overline{\rho}(f) = \max_{e \in E} \tfrac{1}{Q(e)} \sum_{p\ni e}f(p)\mathrm{len}(p),
 \end{align*}
 where $\mathrm{len}(p)$ the length of the path $p$. Then $\lambda_{\max} \le 1 - 1/\overline{\rho}(f)$. 
\end{theorem}

With this property of multicommodity flow, we are ready to prove Thm.~\ref{thm:matroid}.

\begin{proof} \emph{(Theorem~\ref{thm:matroid})}
%
%
%
%
%
We sketch the proof for partition matroids; the full proofs is in Appendix~\ref{app:proofmatroid}. For any two sets $X,Y \in \cc$, we distribute the flow equally across all shortest paths $X \rightsquigarrow Y$ in the transition graph and bound the amount of flow through any edge $e \in E$.

Consider two arbitrary sets $X,Y \in \cc$ with symmetric difference $|X\oplus Y| = 2m\le 2k$, i.e., $m$ elements need to be exchanged to reach from $X$ to $Y$. However, these $m$ steps are a valid path in the transition graph only if every set $S$ along the way is in $\cc$. The exchange property of matroids implies that this requirement is indeed true, so any shortest path  $X \rightsquigarrow Y$ has length $m$. Moreover, there are exactly $m!$ such paths, since we can exchange the elements in $X \setminus Y$ in any order to reach at $Y$. Note that once we choose $s\in X\setminus Y$ to swap out, there is only one choice $t\in Y\setminus X$ to swap in, where $t$ lies in the same part as $s$ in the partition matroid, otherwise the constraint will be violated. Since the total flow is $\prb(X)\prb(Y)$, each path receives $\prb(X)\prb(Y)/m!$ flow.


Next, let $e = (S,T)$ be any edge on some shortest path $X \rightsquigarrow Y$; so $S,T\in\cc$ and $T = S\cup \{j\}\backslash\{i\}$ for some $i, j \in V$. Let $2r = |X\oplus S| < 2m$ be the length of the shortest path $X \rightsquigarrow S$, i.e., $r$ elements need to be exchanged to reach from $X$ to $S$. Similarly, $m-r-1$ elements are exchanged to reach from $T$ to $Y$. Since there is a path for every permutation of those elements, the ratio of the total flow $w_{e}(X,Y)$ that edge $e$ receives from pair $X,Y$, and $Q(e)$, becomes
%
%
\begin{small}
\begin{align} \label{eq:ubound1} 
&{w_e(X,Y)\over Q(e)} \leq  
{2r!(m - 1 - r)! kL\over m!Z_\cc}\exp(2\beta\zeta_F)(\exp(\beta F(\sigma_S(X,Y))) + \exp(\beta F(\sigma_T(X,Y)))),
\end{align}
\end{small}%
where we define $\sigma_{S}(X,Y) = X\oplus Y\oplus S = (X \cap Y \cap S) \cup (X \setminus (Y \cup S)) \cup (Y \setminus (X \cup S))$. 
To bound the total flow, we must count the pairs $X,Y$ such that $e$ is on their shortest path(s), and bound the flow they send. We do this in two steps, first summing over all $(X,Y)$'s that share the upper bound \eqref{eq:ubound1} since they have the same difference sets $U_S = \sigma_{S}(X,Y)$ and $U_T = \sigma_T(X,Y)$, and then we sum over all possible $U_S$ and $U_T$. For fixed $U_S$, $U_T$, there are ${ m-1 \choose r}$ pairs that share those difference sets, since the only freedom we have is to assign $r$ of the $m-1$ elements in $S \setminus (X \cap Y \cap S)$ to $Y$, and the rest to $X$. Hence, for fixed $U_S, U_T$. Appropriate summing and canceling then yields
%
\begin{small}
\begin{align}
\sum_{\substack{ (X,Y):\, \sigma_S(X,Y) = U_S,\\ \sigma_T(X,Y) = U_T}} \frac{w_e(X,Y)}{Q(e)}
\label{eq:USbound1}
&\leq {2kL\over Z_\cc}\exp(2\beta\zeta_F)(\exp(\beta F(U_S)) + \exp(\beta F(U_T))).
\end{align}
\end{small}%
Finally, we sum over all valid $U_S$ ($U_T$ is determined by $U_S$). One can show that any valid $U_S \in \cc$, and hence
$\sum_{U_S} \exp(\beta F(U_S)) \leq Z_\cc$, and likewise for $U_T$. Hence, summing the bound \eqref{eq:USbound1} over all possible choices of $U_S$ yields 
\begin{align*}
\overline{\rho}(f) \le 4kL\exp(2\beta \zeta_F) \max_p \mathrm{len}(p) \leq 4k^2L\exp(2\beta \zeta_F),
\end{align*}
where we upper bound the length of any shortest path by $k$, since $m \leq k$.
Hence
\begin{equation*}
\tau_{X_0} (\eps)\le 4k^2L\exp(2\beta\zeta_F)(\log\pi(X_0)^{-1} + \log\eps^{-1}).\qedhere
\end{equation*}
\end{proof}

For more restrictive constraints, there are fewer paths, and the bounds can become larger. Appendix~\ref{app:proofmatroid} shows the general dependence on $k$ (as $k!$). 
It is also interesting to compare the bound on uniform matroid in Eq.~\eqref{eq:uniformbase} to that shown in~\cite{anari2016monte} for a sub-class of distributions that satisfy the property of being homogeneous strongly Rayleigh\footnote{Appendix~\ref{app:rayleigh} contains details about strongly Rayleigh distributions.}. If $\pi_\cc$ is homogeneous strongly Rayleigh, we have $\tau_{X_0}(\eps)\le 2k(N-k)(\log\pi_\cc(X_0)^{-1} + \log\eps^{-1})$. In our analysis, without additional assumptions on $\pi_\cc$, we pay a factor of $2\exp(2\beta\zeta_F))$ for generality. This factor is one for some strongly Rayleigh distributions (e.g., if $F$ is modular), but not for all. 


\vspace*{-5pt}
\subsection{Uniform Matroid Constraint}
\label{sec:umat}
\vspace*{-5pt}
We consider constraints that is a uniform matroid of certain rank: $\cc = \{S:|S|\le k\}$. We employ the lazy add-delete Markov chain in Algo.~\ref{algo:scmc}, where in each iteration, with probability 0.5 we uniformly randomly sample one element from $V$ and either add it to or delete it from the current set, while respecting constraints. To show fast mixing, we consider using \emph{path coupling}, which essentially says that if we have a contraction of two (coupling) chains then we have fast mixing. We construct path coupling $(S,T)\to (S',T')$ on a carefully generated graph with edges $E$ (from a proper metric). With all details in Appendix~\ref{app:proof:pathcoupling} we end up with the following theorem:

\begin{theorem}\label{thm:size}
Consider the chain shown in Algorithm~\ref{algo:scmc}.
Let $\alpha = \max_{(S,T)\in E}\{\alpha_1,\alpha_2\}$ where $\alpha_1$ and $\alpha_2$ are functions of edges $(S,T)\in E$ and are defined as 
\begin{small}
\begin{align*}
\alpha_1 =& \sum_{i\in T}|p^-(T,i) - p^-(S,i)|_+ + \llbracket |S| < k\rrbracket\sum_{i\in [N]\backslash S}(p^+(S,i) - p^+(T,i))_+;\\
\alpha_2 =& 1 - (\min\{p^-(S,s), p^-(T,t)\} - \sum_{i\in R}|p^-(S,i) - p^-(T,i)| + \\
&\;\;\;\;\llbracket |S| < k\rrbracket (\min\{p^+(S,t), p^+(T,s)\} - \sum_{i\in [N]\backslash (S\cup T)}|p^+(S,i) - p^+(T,i)|)),
\end{align*}
\end{small}%
where $(x)_+ = \max(0,x)$. The summations over absolute differences quantify the sensitivity of transition probabilities to adding/deleting elements in neighboring $(S,T)$. Assuming $\alpha < 1$, we get
\begin{align*}
\tau(\eps) \le {2N\log(N\eps^{-1})\over 1 - \alpha}
\end{align*}
\end{theorem}
\begin{algorithm}
\begin{small}
	\caption{\small Gibbs Add-Delete Markov Chain for Uniform Matroid}\label{algo:scmc}
	\begin{algorithmic} 
	\Require{$F$ the set function, $\beta$ the inverse temperature, $V$ the ground set, $k$ the rank of $\cc$}
	\Ensure{$S$ sampled from $\pi_\cc$}
	\State Initialize $S\in\cc$
	\While{not mixed}
		\State Let $b = 1$ with probability $0.5$
		\If{$b = 1$}
			\State Draw $s\in V$ uniformly randomly
			\If{$s\notin S$ and $|S\cup\{s\}|\le k$}
				\State $S\leftarrow S\cup\{s\}$ with probability $p^+(S,s) = {\pi_\cc(S\cup\{s\})\over \pi_\cc(S) + \pi_\cc(S\cup\{s\})}$
			\Else
				\State $S\leftarrow S\backslash\{s\}$ with probability $p^-(S,s) = {\pi_\cc(S\backslash\{s\})\over \pi_\cc(S) + \pi_\cc(S\backslash\{s\})}$
			\EndIf
		\EndIf
	\EndWhile
\end{algorithmic}
\end{small}
\end{algorithm}

\noindent \textbf{Remarks.} If $\alpha$ is less than 1 and independent of $N$, then the mixing time is nearly linear in $N$. The condition is conceptually similar to those in~\cite{rebeschini2015fast,li2016fast}. The fast mixing requires both $\alpha_1$ and $\alpha_2$, specifically, the change in probability when adding or deleting single element to neighboring subsets, to be small. Such notion is closely related to the \emph{curvature} of discrete set functions. 

\vspace*{-8pt}


\section{Experiments}\label{sec:experiments}

We next empirically study the dependence of sampling times on key factors that govern our theoretical bounds. In particular, we run Markov chains on chain-structured Ising models on a partition matroid base and DPPs on a uniform matroid, and consider estimating marginal and conditional probabilities of a single variable. To monitor the convergence of Markov chains, we use \emph{potential scale reduction factor} (PSRF)~\cite{gelman1992inference,brooks1998general} that runs several chains in parallel and compares within-chain variances to between-chain variances. Typically, PSRF is greater than 1 and will converge to 1 in the limit; if it is close to 1 we empirically conclude that chains have mixed well. Throughout experiments we run 10 chains in parallel for estimations, and declare ``convergence'' at a PSRF of 1.05.

\begin{figure}[h!]
\centering
	\begin{subfigure}{.3\textwidth}
	\centering
	\includegraphics[width=\textwidth]{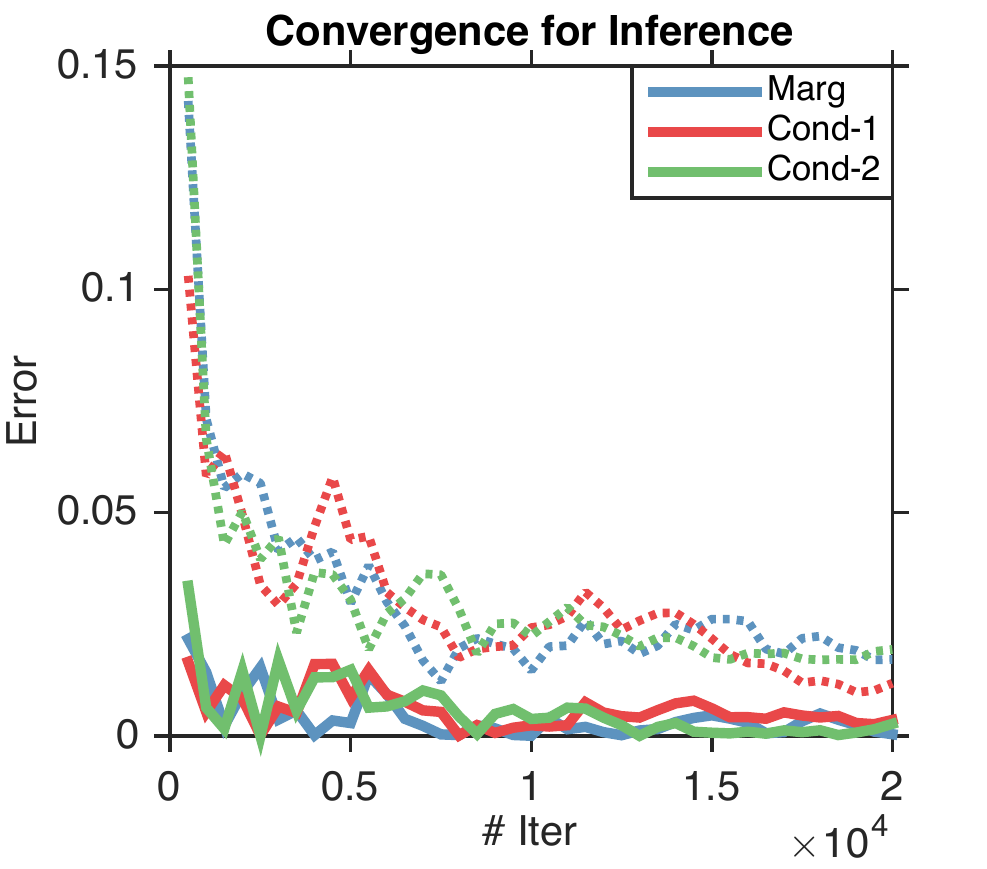}
	\caption{$(\beta,\delta) = (1,1)$}
	\label{fig:ising1}
	\end{subfigure}%
	\begin{subfigure}{.3\textwidth}
	\centering
	\includegraphics[width=\textwidth]{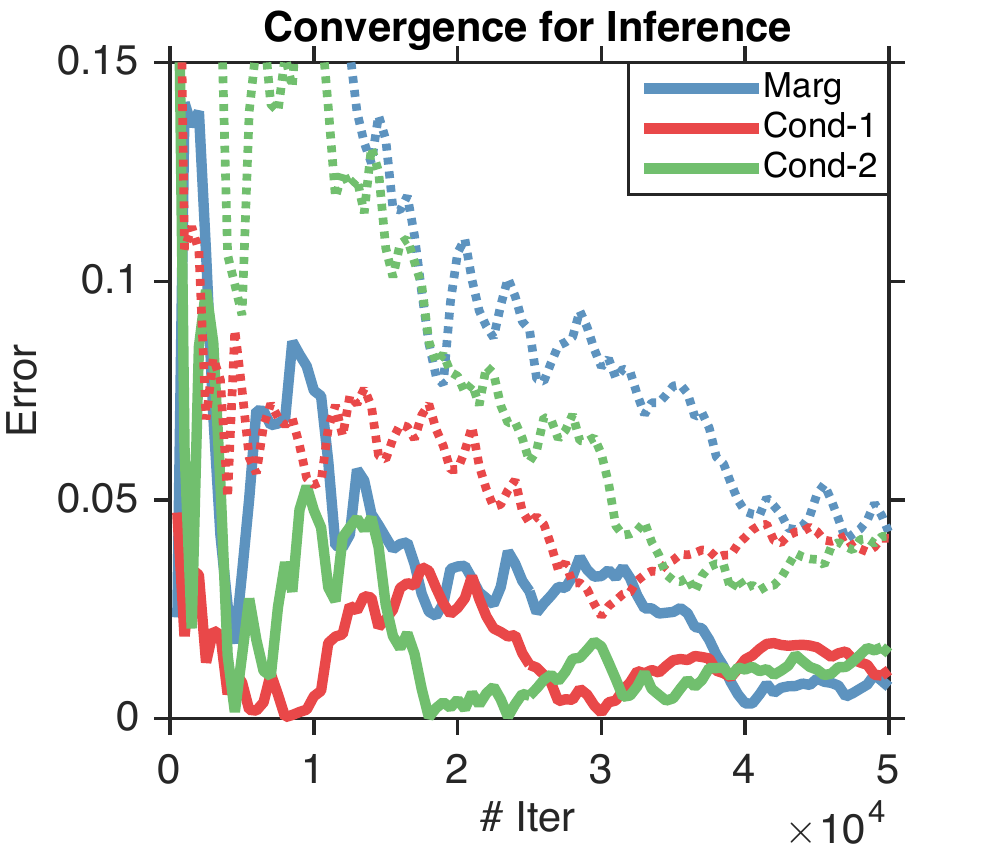}
	\caption{$(\beta,\delta) = (3,1)$}
	\label{fig:ising2}
	\end{subfigure}%
	\begin{subfigure}{.3\textwidth}
	\centering
	\includegraphics[width=\textwidth]{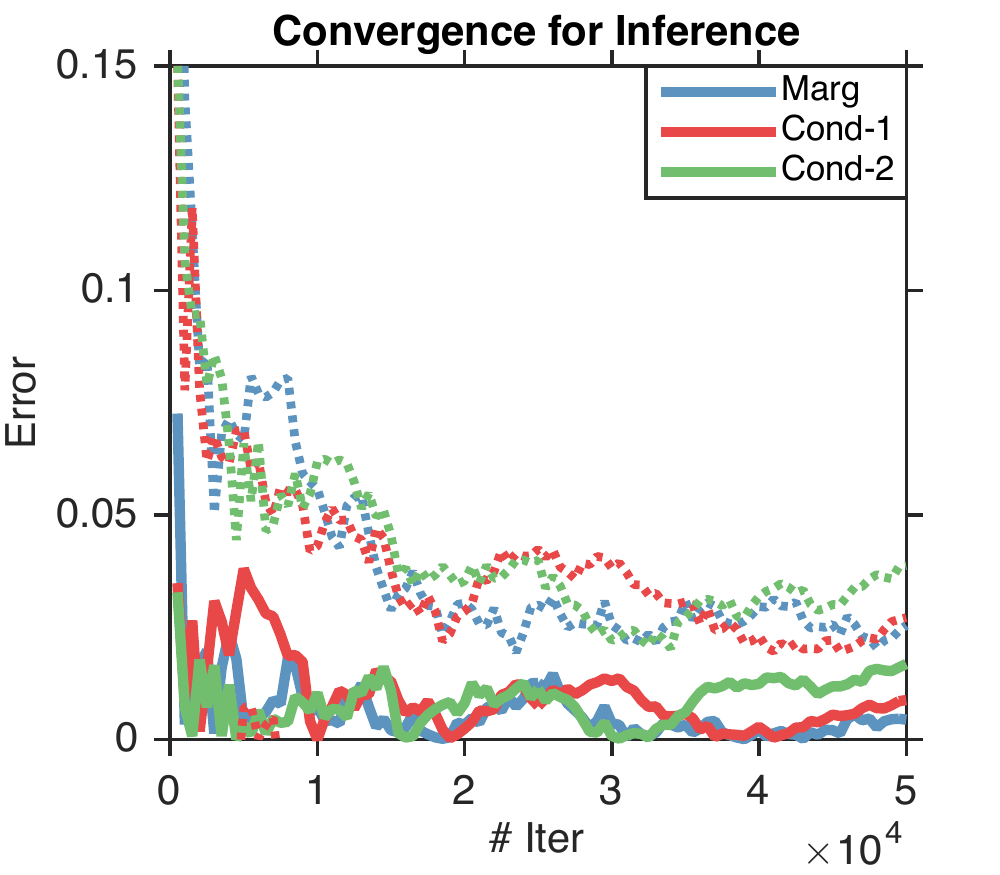}
	\caption{$(\beta,\delta) = (3,0.5)$}
	\label{fig:ising3}
	\end{subfigure}
\caption{Convergence of marginal (\texttt{Marg}) and conditional~(\texttt{Cond-1} and \texttt{Cond-2}, conditioned on $1$ and $2$ other variables) probabilities of a single variable in a 20-variable Ising model with different $(\beta,\delta)$. Full lines show the means and dotted lines the standard deviations of estimations.}
\label{fig:ising}
\end{figure}

We first focus on small synthetic examples where we can compute exact marginal and conditional probabilities. We construct a 20-variable chain-structured Ising model as
\begin{small}
\begin{equation*}
\pi_\cc(S) \propto \exp\left(\beta \left(\left(\delta\sum_{i=1}^{19} w_i (s_i\oplus s_{i+1}) \right) + (1-\delta) |S|\right)\right)\llbracket S\in\cc\rrbracket,
\end{equation*}
\end{small}
where the $s_i$ are 0-1 encodings of $S$, and the $w_i$ are drawn uniformly randomly from $[0,1]$. The parameters $(\beta,\delta)$ govern bounds on the mixing time via $\exp(2\beta\zeta_F)$; the smaller $\delta$, the smaller $\zeta_F$. 
$\cc$ is a partition matroid of rank 5. We estimate conditional probabilities of one random variable conditioned on 0, 1 and 2 other variables and compare against the ground truth. We set $(\beta,\delta)$ to be $(1,1)$, $(3,1)$ and $(3,0.5)$ and results are shown in Fig.~\ref{fig:ising}. All marginals and conditionals converge to their true values, but with different speed. Comparing Fig.~\ref{fig:ising1} against~\ref{fig:ising2}, we observe that with fixed $\delta$, increase in $\beta$ slows down the convergence, as expected. Comparing Fig.~\ref{fig:ising2} against~\ref{fig:ising3}, we observe that with fixed $\beta$, decrease in $\delta$ speeds up the convergence, also as expected given our theoretical results. Appendix~\ref{app:sec:delta} and~\ref{app:sec:beta} illustrate the convergence of estimations under other $(\beta,\delta)$ settings. 

\begin{figure}[h!]
\centering
	\begin{subfigure}{.3\textwidth}
	\centering
	\includegraphics[width=\textwidth]{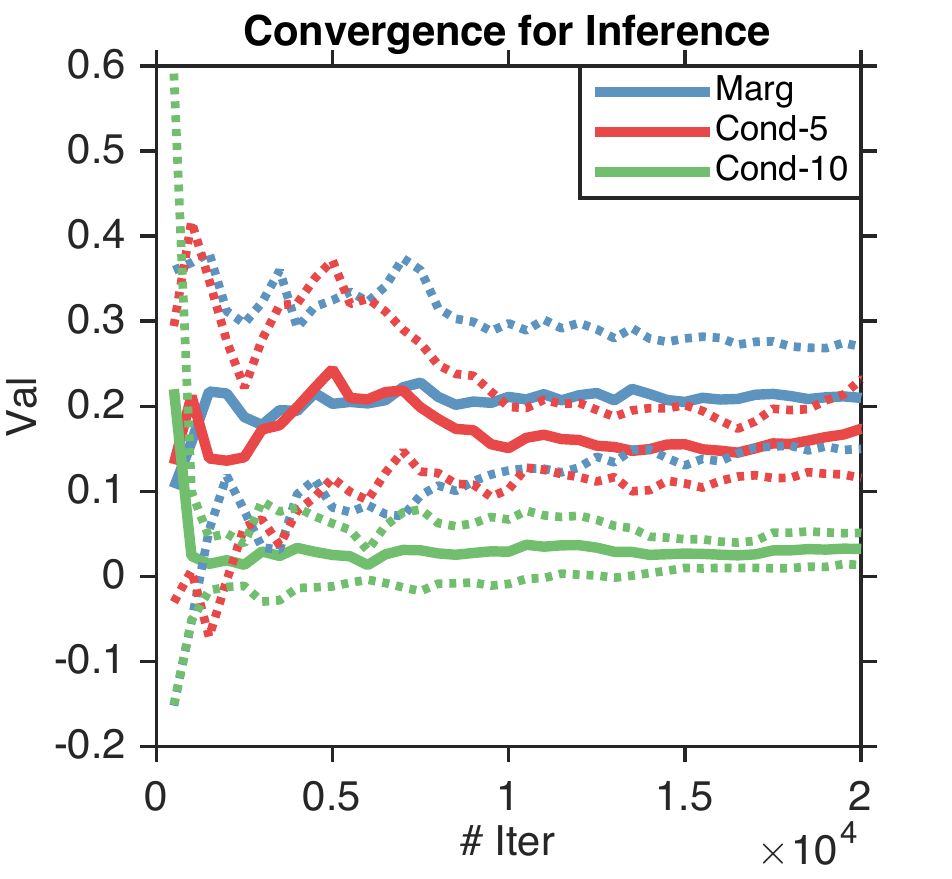}
	\caption{}
	\label{fig:largedpp}
	\end{subfigure}%
	\begin{subfigure}{.32\textwidth}
	\centering
	\includegraphics[width=\textwidth]{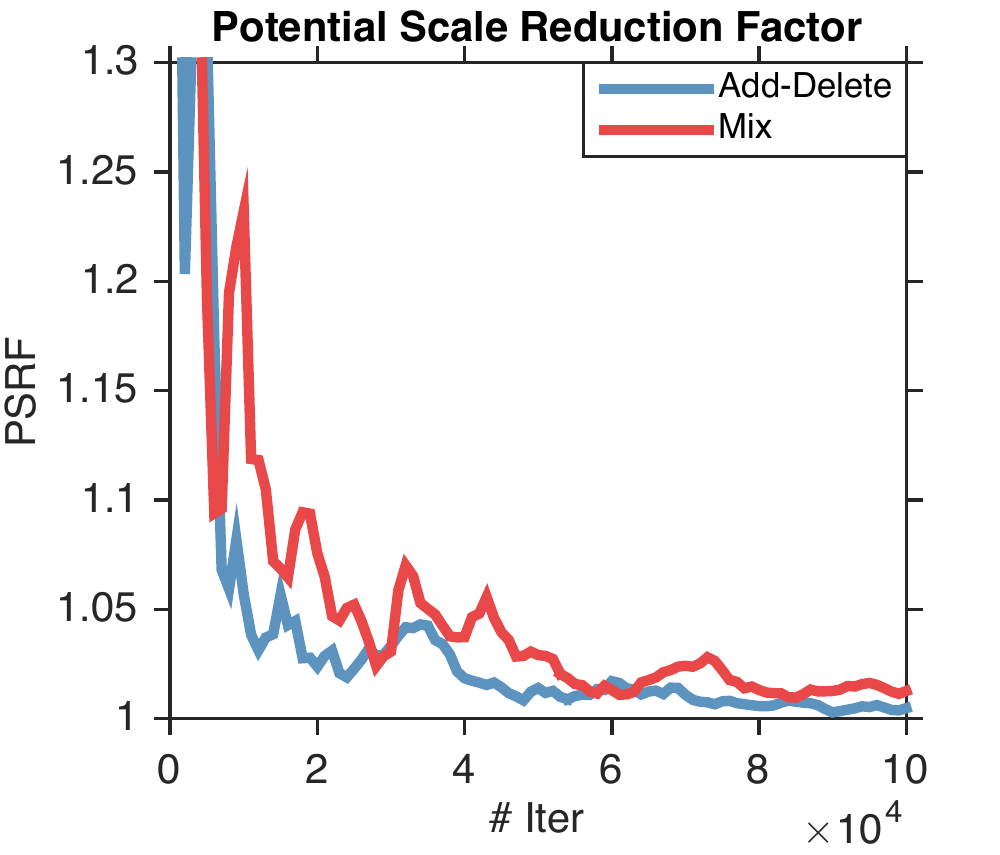}
	\caption{}
	\label{fig:rayleigh1}
	\end{subfigure}%
	\begin{subfigure}{.32\textwidth}
	\centering
	\includegraphics[width=\textwidth]{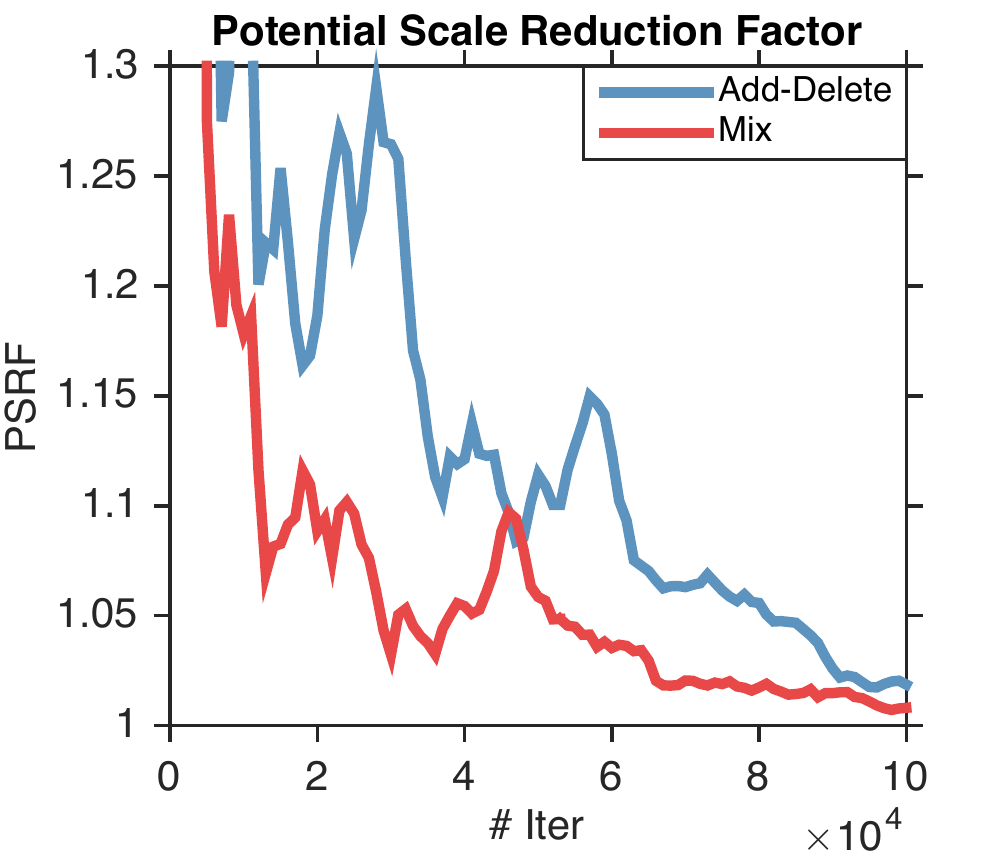}
	\caption{}
	\label{fig:rayleigh2}
	\end{subfigure}
\caption{(a) Convergence of marginal and conditional probabilities by DPP on uniform matroid, (b,c) comparison between add-delete chain (Algorithm~\ref{algo:scmc}) and projection chain (Algorithm~\ref{algo:rayleigh}) for two instances: slowly decaying spectrum and sharp step in the spectrum.}
\label{fig:large}
\end{figure}

We also check convergence on larger models. We use a DPP on a uniform matroid of rank 30 on the Ailerons data ({\url{http://www.dcc.fc.up.pt/ 657\~ltorgo/Regression/DataSets.html}}) of size 200. Here, we do not have access to the ground truth, and hence plot the estimation mean with standard deviations among 10 chains in~\ref{fig:largedpp}. We observe that the chains will eventually converge, i.e., the mean becomes stable and variance small. We also use PSRF to approximately judge the convergence. More results can be found in Appendix~\ref{app:sec:size}. 

\begin{figure}[h!]
\centering
	\begin{subfigure}{.57\textwidth}
	\centering
	\includegraphics[width=\textwidth]{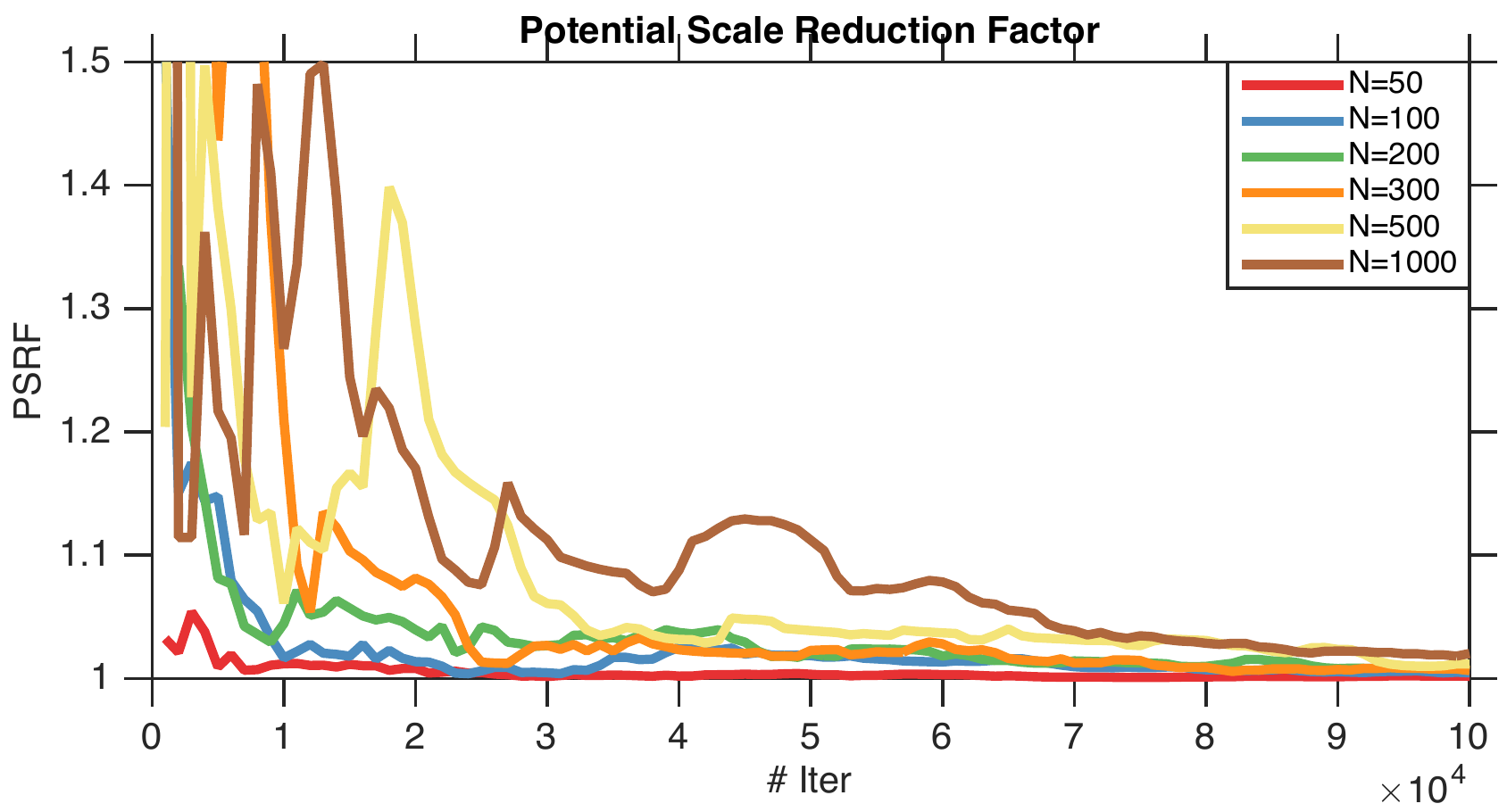}
	\caption{}
	\label{fig:psrfdpp}
	\end{subfigure}%
	\begin{subfigure}{.33\textwidth}
	\centering
	\includegraphics[width=\textwidth]{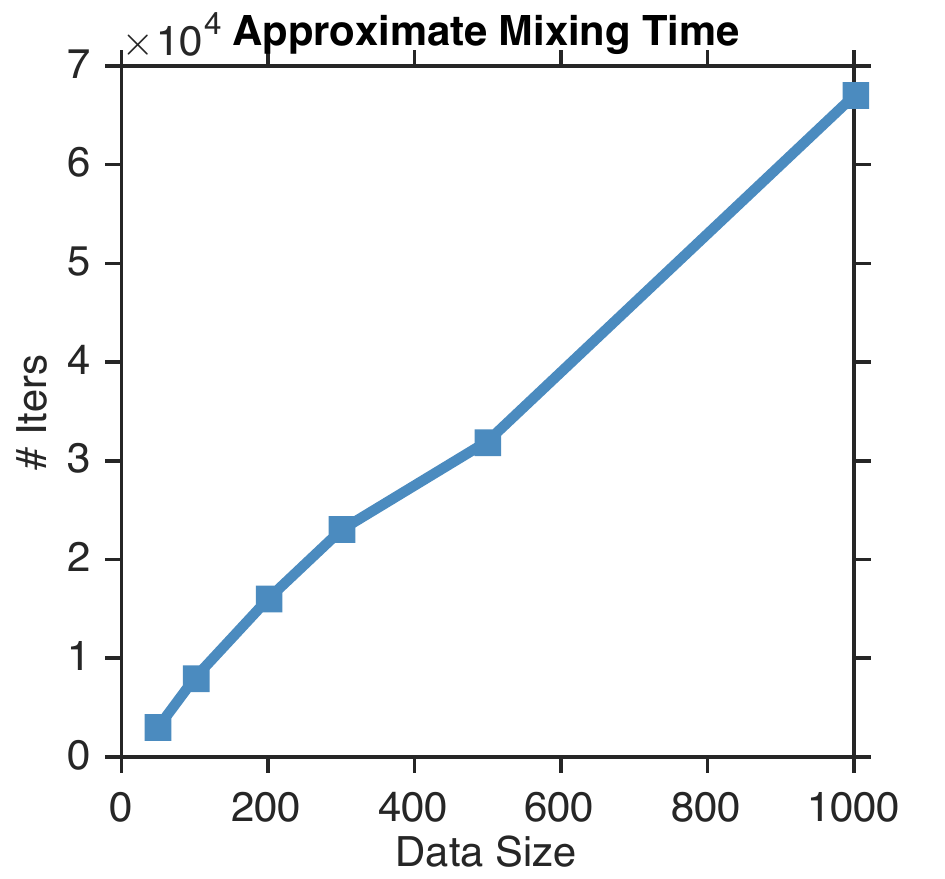}
	\caption{}
	\label{fig:datasize}
	\end{subfigure}
\caption{Empirical mixing time analysis when varying dataset sizes, (a) PSRF's for each set of chains, (b) Approximate mixing time obtained by thresholding PSRF at 1.05.}
\label{fig:convsize}
\end{figure}

Furthermore, the mixing time depends on the size $N$ of the ground set. We use a DPP on Ailerons and vary $N$ from 50 to 1000. Fig.~\ref{fig:psrfdpp} shows the PSRF from 10 chains for each setting. 
By thresholding PSRF at 1.05 in Fig.~\ref{fig:datasize} we see a clearer dependence on $N$. At this scale, the mixing time grows almost linearly with $N$, indicating that this chain is efficient at least at small to medium scale. 
\vspace{-.1in}

Finally, we empirically study how fast our sampler on strongly Rayleigh distribution converges. We compare the chain in Algorithm~\ref{algo:rayleigh}~(\texttt{Mix}) against a simple add-delete chain (\texttt{Add-Delete}). 
We use a DPP on Ailerons data\footnote{\url{http://www.dcc.fc.up.pt/ 657\~ltorgo/Regression/DataSets.html}} of size 200, and the corresponding PSRF is shown in Fig.~\ref{fig:rayleigh1}. We observe that \texttt{Mix} converges slightly slower than \texttt{Add-Delete} since it is lazier. However, the Add-Delete chain does not always mix fast. Fig.~\ref{fig:rayleigh2} illustrates a different setting, where we modify the eigenspectrum of the kernel matrix: the first 100 eigenvalues are 500 and others 1/500. Such a kernel corresponds to almost an elementary DPP, where the size of the observed subsets sharply concentrates around 100. Here,  \texttt{Add-Delete} moves very slowly. \texttt{Mix}, in contrast, has the ability of exchanging elements and thus converges way faster than \texttt{Add-Delete}.

\section{Discussion and Open Problems}
We presented theoretical results on Markov chain sampling for discrete probabilistic models subject to implicit and explicit constraints. In particular, under an implicit constraint that the probability measure is strongly Rayleigh, we obtain an unconditional fast mixing guarantee. For distributions with various explicit constraints we showed sufficient conditions for fast mixing. We show empirically that the dependencies of mixing times on various factors are consistent with our theoretical analysis.

There still exist many open problems in both implicitly- and explicitly-constrained settings. Many bounds that we show depend on structural quantities ($\zeta_F$ or $\alpha$) that may not always be easy to quantify in practice. It will be valuable to develop chains on special classes of distributions (like we did for strongly Rayleigh) whose mixing time is independent of these factors. Moreover, we only considered matroid bases or uniform matroids, while several important settings such as knapsack constraints remain open. In fact, even uniform sampling with a knapsack constraint is not easy; a mixing time of $\co(N^{4.5})$ is known~\cite{morris2004random}. We defer the development of similar or better bounds, potentially with structural factors like $\exp(\beta\zeta_F)$, on specialized discrete probabilistic models as our future work.
\\\\
\noindent\textbf{Acknowledgements.}
This research was partially supported by NSF CAREER 1553284 and a Google Research Award. We thank Ruilin Li for pointing out typos.

\bibliographystyle{abbrvnat}
{  \footnotesize\bibliography{referFinal}}

\begin{thebibliography}{38}
\providecommand{\natexlab}[1]{#1}
\providecommand{\url}[1]{\texttt{#1}}
\expandafter\ifx\csname urlstyle\endcsname\relax
  \providecommand{\doi}[1]{doi: #1}\else
  \providecommand{\doi}{doi: \begingroup \urlstyle{rm}\Url}\fi

\bibitem[Aldous(1982)]{aldous1982some}
D.~J. Aldous.
\newblock Some inequalities for reversible {M}arkov chains.
\newblock \emph{Journal of the London Mathematical Society}, pages 564--576,
  1982.

\bibitem[Anari and Gharan(2015)]{anari15}
N.~Anari and S.~O. Gharan.
\newblock Effective-resistance-reducing flows and asymmetric tsp.
\newblock In \emph{FOCS}, 2015.

\bibitem[Anari et~al.(2016)Anari, Gharan, and Rezaei]{anari2016monte}
N.~Anari, S.~O. Gharan, and A.~Rezaei.
\newblock {M}onte {C}arlo {M}arkov chain algorithms for sampling strongly
  {R}ayleigh distributions and determinantal point processes.
\newblock In \emph{COLT}, 2016.

\bibitem[Borcea et~al.(2009)Borcea, Br{\"a}nd{\'e}n, and
  Liggett]{borcea2009negative}
J.~Borcea, P.~Br{\"a}nd{\'e}n, and T.~Liggett.
\newblock Negative dependence and the geometry of polynomials.
\newblock \emph{Journal of the American Mathematical Society}, pages 521--567,
  2009.

\bibitem[Bouchard-C\^ot\'e and Jordan(2010)]{bouchard10}
A.~Bouchard-C\^ot\'e and M.~I. Jordan.
\newblock Variational inference over combinatorial spaces.
\newblock In \emph{NIPS}, 2010.

\bibitem[Broder(1989)]{broder1989generating}
A.~Broder.
\newblock Generating random spanning trees.
\newblock In \emph{FOCS}, pages 442--447, 1989.

\bibitem[Brooks and Gelman(1998)]{brooks1998general}
S.~P. Brooks and A.~Gelman.
\newblock General methods for monitoring convergence of iterative simulations.
\newblock \emph{Journal of computational and graphical statistics}, pages
  434--455, 1998.

\bibitem[Bubley and Dyer(1997)]{bubley1997path}
R.~Bubley and M.~Dyer.
\newblock Path coupling: A technique for proving rapid mixing in {M}arkov
  chains.
\newblock In \emph{FOCS}, pages 223--231, 1997.

\bibitem[Cesa-Bianchi and Lugosi(2009)]{cesabianchi09}
N.~Cesa-Bianchi and G.~Lugosi.
\newblock Combinatorial bandits.
\newblock In \emph{COLT}, 2009.

\bibitem[Diaconis and Stroock(1991)]{diaconis1991geometric}
P.~Diaconis and D.~Stroock.
\newblock Geometric bounds for eigenvalues of {M}arkov chains.
\newblock \emph{The Annals of Applied Probability}, pages 36--61, 1991.

\bibitem[Djolonga and Krause(2014)]{djolonga2014map}
J.~Djolonga and A.~Krause.
\newblock From {MAP} to marginals: Variational inference in bayesian submodular
  models.
\newblock In \emph{NIPS}, pages 244--252, 2014.

\bibitem[Dyer and Greenhill(1998)]{dyer1998more}
M.~Dyer and C.~Greenhill.
\newblock A more rapidly mixing {M}arkov chain for graph colorings.
\newblock \emph{Random Structures and Algorithms}, pages 285--317, 1998.

\bibitem[Dyer et~al.(1999)Dyer, Frieze, and Jerrum]{dyer99}
M.~Dyer, A.~Frieze, and M.~Jerrum.
\newblock On counting independent sets in sparse graphs.
\newblock In \emph{FOCS}, 1999.

\bibitem[Ermon et~al.(2013)Ermon, Gomes, Sabharwal, and Selman]{ermon2013embed}
S.~Ermon, C.~P. Gomes, A.~Sabharwal, and B.~Selman.
\newblock Embed and project: Discrete sampling with universal hashing.
\newblock In \emph{NIPS}, pages 2085--2093, 2013.

\bibitem[Feder and Mihail(1992)]{feder1992balanced}
T.~Feder and M.~Mihail.
\newblock Balanced matroids.
\newblock In \emph{STOC}, pages 26--38, 1992.

\bibitem[Frieze et~al.(2014)Frieze, Goyal, Rademacher, and Vempala]{frieze14}
A.~Frieze, N.~Goyal, L.~Rademacher, and S.~Vempala.
\newblock Expanders via random spanning trees.
\newblock \emph{SIAM Journal on Computing}, 43\penalty0 (2):\penalty0 497--513,
  2014.

\bibitem[Gartrell et~al.(2016)Gartrell, Paquet, and Koenigstein]{gartrell16}
M.~Gartrell, U.~Paquet, and N.~Koenigstein.
\newblock Low-rank factorization of determinantal point processes for
  recommendation.
\newblock \emph{arXiv:1602.05436}, 2016.

\bibitem[Gelman and Rubin(1992)]{gelman1992inference}
A.~Gelman and D.~B. Rubin.
\newblock Inference from iterative simulation using multiple sequences.
\newblock \emph{Statistical science}, pages 457--472, 1992.

\bibitem[Gotovos et~al.(2015)Gotovos, Hassani, and Krause]{gotovos2015sampling}
A.~Gotovos, H.~Hassani, and A.~Krause.
\newblock Sampling from probabilistic submodular models.
\newblock In \emph{NIPS}, 2015.

\bibitem[Greig et~al.(1989)Greig, Porteous, and Seheult]{gps89}
D.~M. Greig, B.~T. Porteous, and A.~H. Seheult.
\newblock Exact maximum a posteriori estimation for binary images.
\newblock \emph{Journal of the Royal Statistical Society}, 1989.

\bibitem[Iyer and Bilmes(2015)]{iyer2015submodular}
R.~Iyer and J.~Bilmes.
\newblock Submodular point processes.
\newblock In \emph{AISTATS}, 2015.

\bibitem[Jerrum and Sinclair(1993)]{jerrum93ising}
M.~Jerrum and A.~Sinclair.
\newblock Polynomial-time approximation algorithms for the {I}sing model.
\newblock \emph{SIAM J. Computing}, 1993.

\bibitem[Jerrum et~al.(2004)Jerrum, Sinclair, and Vigoda]{jerrum04}
M.~Jerrum, A.~Sinclair, and E.~Vigoda.
\newblock A polynomial-time approximation algorithm for the permanent of a
  matrix with nonnegative entries.
\newblock \emph{JACM}, 2004.

\bibitem[Kang(2013)]{kang2013fast}
B.~Kang.
\newblock Fast determinantal point process sampling with application to
  clustering.
\newblock In \emph{NIPS}, pages 2319--2327, 2013.

\bibitem[Kathuria and Deshpande(2016)]{kathuria2016sampling}
T.~Kathuria and A.~Deshpande.
\newblock On sampling from constrained diversity promoting point processes.
\newblock 2016.

\bibitem[Kojima and Komaki(2014)]{kojima2014determinantal}
M.~Kojima and F.~Komaki.
\newblock Determinantal point process priors for {B}ayesian variable selection
  in linear regression.
\newblock \emph{arXiv:1406.2100}, 2014.

\bibitem[Kulesza and Taskar(2011)]{kulesza2011k}
A.~Kulesza and B.~Taskar.
\newblock k-{DPP}s: Fixed-size determinantal point processes.
\newblock In \emph{ICML}, pages 1193--1200, 2011.

\bibitem[Kulesza and Taskar(2012)]{kulesza2012determinantal}
A.~Kulesza and B.~Taskar.
\newblock Determinantal point processes for machine learning.
\newblock \emph{arXiv preprint arXiv:1207.6083}, 2012.

\bibitem[Li et~al.(2016{\natexlab{a}})Li, Jegelka, and Sra]{li2016fast}
C.~Li, S.~Jegelka, and S.~Sra.
\newblock Fast {DPP} sampling for {N}ystr{\"{o}}m with application to kernel
  methods.
\newblock In \emph{ICML}, 2016{\natexlab{a}}.

\bibitem[Li et~al.(2016{\natexlab{b}})Li, Sra, and Jegelka]{li2016gauss}
C.~Li, S.~Sra, and S.~Jegelka.
\newblock Gaussian quadrature for matrix inverse forms with applications.
\newblock In \emph{ICML}, 2016{\natexlab{b}}.

\bibitem[Maddison et~al.(2014)Maddison, Tarlow, and Minka]{a2013maddison}
C.~J. Maddison, D.~Tarlow, and T.~Minka.
\newblock A* sampling.
\newblock In \emph{NIPS}, 2014.

\bibitem[Mariet and Sra(2016)]{mariet16}
Z.~Mariet and S.~Sra.
\newblock Diversity networks.
\newblock In \emph{ICLR}, 2016.

\bibitem[Morris and Sinclair(2004)]{morris2004random}
B.~Morris and A.~Sinclair.
\newblock Random walks on truncated cubes and sampling 0-1 knapsack solutions.
\newblock \emph{SIAM journal on computing}, pages 195--226, 2004.

\bibitem[Rebeschini and Karbasi(2015)]{rebeschini2015fast}
P.~Rebeschini and A.~Karbasi.
\newblock Fast mixing for discrete point processes.
\newblock In \emph{COLT}, 2015.

\bibitem[Sinclair(1992)]{sinclair1992improved}
A.~Sinclair.
\newblock Improved bounds for mixing rates of {M}arkov chains and
  multicommodity flow.
\newblock \emph{Combinatorics, probability and Computing}, pages 351--370,
  1992.

\bibitem[Smith and Eisner(2008)]{smith08}
D.~Smith and J.~Eisner.
\newblock Dependency parsing by belief propagation.
\newblock In \emph{EMNLP}, 2008.

\bibitem[Spielman and Srivastava(2008)]{spielman08}
D.~Spielman and N.~Srivastava.
\newblock Graph sparsification by effective resistances.
\newblock In \emph{STOC}, 2008.

\bibitem[Zhang et~al.(2015)Zhang, Djolonga, and Krause]{zhang2015higher}
J.~Zhang, J.~Djolonga, and A.~Krause.
\newblock Higher-order inference for multi-class log-supermodular models.
\newblock In \emph{ICCV}, pages 1859--1867, 2015.

\end{thebibliography}

\newpage

\begin{appendix}

\section{Proof of Thm.~\ref{thm:matroid}}\label{app:proofmatroid}

\subsection{Proof for Uniform Matroid Base}
\begin{proof}
We consider the case where $\cc$ is uniform matroid base. For any two sets $X,Y \in \cc$, we distribute the flow equally across all shortest paths $X \rightsquigarrow Y$ in the transition graph. Then, for arbitrary edge $e \in E$, we bound the number of paths (and flow) through $e$.

Consider two arbitrary sets $X,Y \in \cc$ with symmetric difference $|X\oplus Y| = 2m\le 2k$. Any shortest path  $X \rightsquigarrow Y$ has length $m$. Moreover, there are exactly $(m!)^2$ such paths, since we can exchange the elements in $X \setminus Y$ in any order with the elements in $Y\setminus X$ in any order to reach at $Y$. Since the total flow is $\prb(X)\prb(Y)$, each path receives $\prb(X)\prb(Y)/(m!)^2$ flow.

Next, let $e = (S,T)$ be any edge on some shortest path $X \rightsquigarrow Y$; so $S,T\in\cc$ and $T = S\cup \{j\}\backslash\{i\}$ for some $i, j \in [N]$. Let $2r = |X\oplus S| < 2m$ be the length of the shortest path $X \rightsquigarrow S$, thus there are $(r!)^2$ ways to reach from $X$ to $S$. Similarly, $m-r-1$ elements are exchanged to reach from $T$ to $Y$ and there are in total $((m-r-1)!)^2$ ways to do so. the total flow $e$ receives from pair $X,Y$ is
\begin{align*}
w_{e}(X,Y) = {\prb(X)\prb(Y)\over (m!)^2}(r!)^2((m - 1 - r)!)^2
\end{align*}
Since in our chain,
\begin{align*}
Q(e) &= {2Z_\cc\exp(\beta F(S))\exp(\beta F(T))\over k(N-k)(\exp(\beta F(S))+ \exp(\beta F(T)))},
\end{align*}
it follows that
\begin{align*}
{w_e(X,Y)\over Q(e)} &=  {2(r!)^2((m - 1 - r)!)^2 k(N-k)\exp(\beta(F(X) + F(Y)))(\exp(\beta F(S)) + \exp(\beta F(T)))\over (m!)^2Z_\cc\exp(\beta(F(S) + F(T)))}\\
&\le {2(r!)^2((m - 1 - r)!)^2 k(N-k)\over (m!)^2Z_\cc}\exp(2\beta\zeta_F)(\exp(\beta F(\sigma_S(X,Y))) + \exp(\beta F(\sigma_T(X,Y)))),
\end{align*}
where we define $\sigma_{S}(X,Y) = X\oplus Y\oplus S$. The inequality draws from the fact that
\begin{align*}
&{\exp(\beta(F(X) + F(Y) + F(S)))\over \exp(\beta(F(S) + F(T)))} = \exp(\beta (F(X) + F(Y) - F(T))\\
&= \exp(\beta(F(X) + F(Y) - F(X\cap Y) - F(X\cup Y)))\\
&\;\; \exp(\beta(F(X\cap Y)+ F(X\cup Y) - F(T) - F(\sigma_T(X,Y))))\exp(\beta F(\sigma_T(X,Y))\\
&\le \exp(2\beta \zeta_F) \exp(\beta F(\sigma_T(X,Y)))
\end{align*}
and likewise for ${\exp(\beta(F(X) + F(Y) + F(T)))\over \exp(\beta(F(S) + F(T)))}$. Similar trick has been used in~\cite{gotovos2015sampling}. 

Let $U_S = \sigma_{S}(X,Y)$ and $U_T = \sigma_T(X,Y)$, then for fixed $U_S, U_T$, the total flow that passes $e$ is
\begin{align*}
&\sum_{\substack{ (X,Y):\, \sigma_S(X,Y) = U_S,\\ \sigma_T(X,Y) = U_T}} {w_e(X,Y)\over Q(e)}\\
&\le 2\sum_{r=0}^{m-1} {m-1\choose r}^2{(r!)^2((m - 1 - r)!)^2 k(N-k)\over (m!)^2Z}\\ 
&\quad\quad\quad\quad\quad\quad\times\exp(2\beta\zeta_F)(\exp(\beta F(U_S)) + \exp(\beta F(U_T)))\\
&= {2k(N-k)\over mZ_\cc}\exp(2\beta\zeta_F)(\exp(\beta F(U_S)) + \exp(\beta F(U_T))).
\end{align*}
Finally, with the definition of $\overline{\rho}(f)$ we sum over all images of $U_S$ and $U_T$. Recall that $Z = \sum_{U_S}\exp(\beta F(U_S))$. Since $|S\oplus X\oplus Y| = k$ we know that $U_S, U_T\in \cc$, thus $Z\le Z_\cc$ and 
\begin{align*}
\overline{\rho}(f) \le 4k(N-k)\exp(2\beta \zeta_F).
\end{align*}
Hence
\begin{equation*}
\tau_{X_0} (\eps)\le 4k(N-k)\exp(2\beta\zeta_F)(\log\prb(X_0)^{-1} + \log\eps^{-1}).
\end{equation*}

\subsection{Proof on Partition Matroid Base}

\begin{proof}
Consider two arbitrary sets $X,Y \in \cc$ with symmetric difference $|X\oplus Y| = 2m\le 2k$, i.e., $m$ elements need to be exchanged to reach from $X$ to $Y$. However, these $m$ steps are a valid path in the transition graph only if every set $S$ along the way is in $\cc$. The exchange property of matroids implies that this is indeed true, so any shortest path  $X \rightsquigarrow Y$ has length $m$. Moreover, there are exactly $m!$ such paths, since we can exchange the elements in $X \setminus Y$ in any order to reach at $Y$. Note that once we choose $s\in X\setminus Y$ to swap out, there is only one choice $t\in Y\setminus X$ to swap in, where $t$ lies in the same part as $s$ in the partition matroid, otherwise the constraint will be violated. Since the total flow is $\prb(X)\prb(Y)$, each path receives $\prb(X)\prb(Y)/m!$ flow.

Next, let $e = (S,T)$ be any edge on some shortest path $X \rightsquigarrow Y$; so $S,T\in\cc$ and $T = S\cup \{j\}\backslash\{i\}$ for some $i, j \in V$. Let $2r = |X\oplus S| < 2m$ be the length of the shortest path $X \rightsquigarrow S$, i.e., $r$ elements need to be exchanged to reach from $X$ to $S$. Similarly, $m-r-1$ elements are exchanged to reach from $T$ to $Y$. Since there is a path for every permutation of those elements, the total flow edge $e$ receives from pair $X,Y$ is

\begin{align*}
w_{e}(X,Y) = {\prb(X)\prb(Y)\over m!}r!(m - 1 - r)!.
\end{align*}
Since, in our chain, (using $L = \max_i|\cp_i|-1$) 
\begin{align*}
Q(e) &\geq  {\prb(S) \over 2 k L} {\prb(T)\over \prb(S) + \prb(T)} = {\exp(\beta F(S))\exp(\beta F(T))\over 2kLZ_\cc(\exp(\beta F(S))+ \exp(\beta F(T)))},
\end{align*}
it follows that
\begin{align} \nonumber
&{w_e(X,Y)\over Q(e)} \leq  {2r!(m - 1 - r)! kL\exp(\beta(F(X) + F(Y)))(\exp(\beta F(S)) + \exp(\beta F(T)))\over m!Z_\cc\exp(\beta(F(S) + F(T)))}\\
\label{eq:ubound}
&\ \ \ \ \ \ \ \ \le {2r!(m - 1 - r)! kL\over m!Z_\cc}\exp(2\beta\zeta_F)(\exp(\beta F(\sigma_S(X,Y))) + \exp(\beta F(\sigma_T(X,Y)))),
\end{align}
where we define $\sigma_{S}(X,Y) = X\oplus Y\oplus S = (X \cap Y \cap S) \cup (X \setminus (Y \cup S)) \cup (Y \setminus (X \cup S))$. To bound the total flow, we must count the pairs $X,Y$ such that $e$ is on their shortest path(s), and bound the flow they send. We do this in two steps, first summing over all $X,Y$ that share the upper bound \eqref{eq:ubound} since they have the same difference sets $U_S = \sigma_{S}(X,Y)$ and $U_T = \sigma_T(X,Y)$, and then we sum over all possible $U_S$ and $U_T$. For fixed $U_S$, $U_T$, there are ${ m-1 \choose r}$ pairs that share those difference sets, since the only freedom we have is to assign $r$ of the $m-1$ elements in $S \setminus (X \cap Y \cap S)$ to $Y$, and the rest to $X$. Hence, for fixed $U_S, U_T$:

\begin{align}
  \nonumber
\sum_{\substack{ (X,Y):\, \sigma_S(X,Y) = U_S,\\ \sigma_T(X,Y) = U_T}} \frac{w_e(X,Y)}{Q(e)}
\nonumber
&\le 2\sum_{r=0}^{m-1} {m-1\choose r}{r!(m - 1 - r)! kL\over m! Z_\cc}\\ \nonumber
&\quad\quad\quad\quad\quad\quad\times\exp(2\beta\zeta_F)(\exp(\beta F(U_S)) + \exp(\beta F(U_T)))\\
\label{eq:USbound}
&= {2kL\over Z_\cc}\exp(2\beta\zeta_F)(\exp(\beta F(U_S)) + \exp(\beta F(U_T))).
\end{align}
Finally, we sum over all valid $U_S$ ($U_T$ is determined by $U_S$), 
where by ``valid'' we mean there exists $X,Y\in\cc$ and $S\in \cc$ on one path from $X$ to $Y$ such that, $U_S = \sigma_S(X,Y)$. Any such $U_S$ can be constructed by picking $k-m$ elements from $S$ (including $i$), and by replacing the remaining elements $u \in S$ by another member of their partition: i.e., if $u \in \cp_\ell$, then it is replaced by some other $v \in \cp_\ell$, since both $X$ and $Y$ must be in $\cc$. Hence, any $U_S$ satisfies the partition constraint, i.e., $U_S \in \cc$ and therefore $\sum_{U_S} \exp(\beta F(U_S)) \leq Z_\cc$, and likewise for $U_T$. Hence, summing the bound \eqref{eq:USbound} over all possible $U_S$ yields 
\begin{align*}
\overline{\rho}(f) \le 4kL\exp(2\beta \zeta_F) \max_p \mathrm{len}(p) \leq 4k^2L\exp(2\beta \zeta_F),
\end{align*}
where we upper bound the length of any shortest path by $k$, since $m \leq k$.
Hence
\begin{equation*}
\tau_{X_0} (\eps)\le 4k^2L\exp(2\beta\zeta_F)(\log\pi_\cc(X_0)^{-1} + \log\eps^{-1}).\qedhere
\end{equation*}
\end{proof}

\subsection{Proof for General Matroid Base}
In the case where no structural assumption is made on $\cc$, the proof needs to be more carefully handled. Because in this case, we know neither the number of legal paths between any two states, nor the number of $\sigma_S(X,Y)$ falls out of $\cc$. 

We again consider arbitrary sets $X,Y\in \cc$ where $|X\oplus Y| = 2m\le 2k$. The total number of shortest paths is \emph{at least} $(m!)$ due to exchange property of matroids. Since the amount of flow from $X$ to $Y$ is $\prb(X)\prb(Y)$, each path receives \emph{at most} $\prb(x)\prb(y) / m!$ . 

Next, let $e = (S,T)$ be any edge on some shortest path $X \rightsquigarrow Y$; so $S,T\in\cc$ and $T = S\cup \{j\}\backslash\{i\}$ for some $i, j \in V$. Let $2r = |X\oplus S| < 2m$ be the length of the shortest path $X \rightsquigarrow S$, thus there are at most $(r!)^2$ ways to reach from $X$ to $S$. Likewise there are at most $((m-r-1)!)^2$ paths to reach from $T$ to $Y$. The total flow edge $e$ receives from pair $X,Y$ is then upper-bounded as
\begin{align*}
w_{e}(X,Y) \le {\prb(X)\prb(Y)\over m!}(r!)^2((m - 1 - r)!)^2.
\end{align*}
It follows that
\begin{align*}
{w_e(X,Y)\over Q(e)} & \le {2(r!)^2((m - 1 - r)!)^2 k(N-k)\over m!Z_\cc}\exp(2\beta\zeta_F)(\exp(\beta F(U_S)) + \exp(\beta F(U_T))).
\end{align*}
The total pairs of $(X,Y)$ that passes $e$ with the same set of images is upper-bounded by ${m-1\choose r}^2$, thus the flow that passes $e$ with the same set of images is bounded as
\begin{align*}
&\sum_{\substack{ (X,Y):\, \sigma_S(X,Y) = U_S,\\ \sigma_T(X,Y) = U_T}} {w_e(X,Y)\over Q(e)}\\
&\le 2\sum_{r=0}^{m-1} {m-1\choose r}^2{(r!)^2((m - 1 - r)!)^2 k(N-k)\over m!Z}\\ 
&\quad\quad\quad\quad\quad\quad\times\exp(2\beta\zeta_F)(\exp(\beta F(U_S)) + \exp(\beta F(U_T)))\\
&= {2(m-1)!k(N-k)\over Z_\cc}\exp(2\beta\zeta_F)(\exp(\beta F(U_S)) + \exp(\beta F(U_T))).
\end{align*}
Thus if we sum over all $U_S,U_T$, the result is upper-bounded as
\begin{align*}
\overline{\rho}(f) & \le {4k! Z\over Z_\cc}k(N-k)\exp(2\beta \zeta_F).
\end{align*}
Note that here we upper-bounded $m$ with $k$ and $Z$ could be larger than $Z_\cc$ because it may happen that $U_S\notin \cc$. It follows that
\begin{equation*}
\tau_{X_0} (\eps)\le {4k! Z\over Z_\cc}k(N-k)\exp(2\beta\zeta_F)(\log\prb(X_0)^{-1} + \log\eps^{-1}). \qedhere
\end{equation*}
\end{proof}

\section{Proof of Thm.~\ref{thm:size}}\label{app:proof:pathcoupling}

Assume we have a chain $(X_t)$ on state space $V$ with transition matrix $P$, a \emph{coupling} is a new chain $(X_t,Y_t)$ on $V\times V$ such that both $(X_t)$ and $(Y_t)$, if considered marginally, are Markov chains with the same transition matrices $P$. The key point of coupling is to construct such a new chain to encourage $X_t$ and $Y_t$ to \emph{coalesce} quickly. If, in the new chain, $\Pr(X_t\ne Y_t)\le \eps$ for some fixed $t$ regardless of the starting state $(X_0,Y_0)$, then $\tau(\eps)\le t$~\cite{aldous1982some}. To make the coupling construction easier, \emph{Path coupling}~\cite{bubley1997path} is then introduced so as to reduce the coupling to adjacent states in an appropriately constructed state graph. The coupling of arbitrary states follows by aggregation over a path between the two. Path coupling is formalized in the following lemma.

\begin{lemma} \cite{bubley1997path,dyer1998more}
\label{lem:pathcoupling}
Let $\delta$ be an integer-valued metric on $V\times V$ where $\delta(\cdot,\cdot)\le D$. Let $E$ be a subset of $V\times V$ such that for all $(X_t,Y_t)\in V\times V$ there exists a path $X_t = Z^0,\ldots, Z^r = Y_t$ between $X_t$ and $Y_t$ where $(Z^i,Z^{i+1})\in E$ for $i\in[r-1]$ and $\sum_i \delta(Z^i,Z^{i+1}) = \delta(X_t,Y_t)$. Suppose a coupling $(S,T)\to(S',T')$ of the Markov chain is defined on all pairs in $E$ such that there exists an $\alpha < 1$ such that $\mathbb{E}[\delta(S',T')]\le \alpha \delta(S,T)$ for all $(S,T)\in E$, then we have $\tau(\eps)\le {\log(D\eps^{-1})\over (1 - \alpha)}$.
\end{lemma}

We now are ready to state our proof.

\begin{proof}
We define $\delta(X,Y) = {1\over 2}(|X\oplus Y| + ||X|-|Y||)$. It is clear that $\delta(X,Y)\ge 1$ for $X\ne Y$. Let $E = \{(X,Y):\delta(X,Y)= 1\}$ be the set of adjacent states (neighbors), and it follows that $\delta(\cdot,\cdot)$ is a metric satisfying conditions in Lemma~\ref{lem:pathcoupling}. Also we have $\delta(X,Y)\le k$.

We consider constructing a path coupling between any two states $S$ and $T$ with $\delta(S,T) = 1$, $S'$ and $T'$ be the two states after transition. We sample $c_S, c_T\in\{0,1\}$, if $c_S$ is 0 then $S'=S$ and the same with $c_T$. $i_S, i_T\in V$ are drawn uniformly randomly. We consider two possible settings for $S$ and $T$:
\begin{enumerate}
\item If $S$ or $T$ is a subset of the other, we assume without of generality that $S = T\cup \{t\}$. In this setting we always let $i_S = i_T = i$. Then 
	\begin{enumerate}
	\item If $i = t$, we let $c_S = 1 - c_T$;
		\begin{enumerate}
		\item If $c_S = 1$ then $\delta(S',T') = 0$ with probability $p^-(S,t)$;
		\item If $c_S = 0$ then $\delta(S',T') = 0$ with probability $p^+(T,t)$;
		\end{enumerate}
	\item If $i\in T$, we set $c_S = c_T$;
		\begin{enumerate}
		\item If $c_S = 1$ then $\delta(S',T') = 2$ with probability $(p^-(T,i) - p^-(S,i))_+$;
		\end{enumerate}
	\item If $i\in V\backslash S$, we set $c_S = c_T$;
		\begin{enumerate}
		\item If $c_S = 1$ and $|S| < k$ then $\delta(S',T') = 2$ with probability $(p^+(S,i) - p^+(T,i))_+$.
		\end{enumerate}
	\end{enumerate}
	
\item If $S$ and $T$ are of the same sizes, let $S = R\cup \{s\}$ and $T = R\cup \{t\}$. In this setting we always let $c_S = c_T = c$. We consider the case of $c = 1$:
	\begin{enumerate}
	\item If $i_S = s$, let $i_T = t$. Then $\delta(S',T') = 0$ with probability $\min\{p^-(S,s), p^-(T,t)\}$;
	\item If $i_S = t$, let $i_T = s$. If $|S| < k$, Then $\delta(S',T') = 0$ with probability $\min\{p^+(S,t), p^+(T,s)\}$;
	\item If $i_S \in R$, let $i_T = i_S$. Then $\delta(S',T') = 2$ with probability $|p^-(S,i_S) - p^-(T,i_T)|$;
	\item If $i_S \in V\backslash (S\cup T)$, let $i_T = i_S$. If $|S| < k$, Then $\delta(S',T') = 2$ with probability $|p^+(S,i_S) - p^+(T,i_T)|$.
	\end{enumerate}
\end{enumerate}
In all cases where we didn't specify $\delta(S',T')$, it will be $\delta(S',T') = 1$. In the first case of $S = T\cup \{t\}$ we have
\begin{align*}
{\mathbb{E}[\delta(S',T')]\over \mathbb{E}[\delta(S,T)]} \le &{1\over 2N}( (1-p^-(S,t)) + (1-p^+(T,t)) + (2|T|+\sum_{i\in T}(p^-(T,i) - p^-(S,i))_+) + \\
&\;(2(N-|S|) + \llbracket |S| < k\rrbracket\sum_{i\in [N]\backslash S}(p^+(S,i) - p^+(T,i))_+ ))\\
= 1 -   {1\over 2N}(1 - &\sum_{i\in T}(p^-(T,i) - p^-(S,i))_+ - \llbracket |S| < k\rrbracket\sum_{i\in [N]\backslash S}(p^+(S,i) - p^+(T,i))_+) = 1 - {1 - \alpha_1 \over 2N},
\end{align*}
while in the second case of $|S| = R\cup\{s\}$ and $T = R\cup\{t\}$ we have
\begin{align*}
{\mathbb{E}[\delta(S',T')]\over \mathbb{E}[\delta(S,T)]} \le &{1\over 2N}( (1-\min\{p^-(S,s), p^-(T,t)\}) + (1-\llbracket |S| < k\rrbracket\min\{p^+(S,t), p^+(T,s)\}) + \\
&(2|R|+\sum_{i\in R}|p^-(S,i) - p^-(T,i)|) + \\
&(2(N-|S|-1) + \llbracket |S| < k\rrbracket \sum_{i\in [N]\backslash (S\cup T)}|p^+(S,i) - p^+(T,i)|) )\\
= 1 - {1\over 2N}(&\min\{p^-(S,s), p^-(T,t)\} - \sum_{i\in R}|p^-(S,i) - p^-(T,i)| + \\
\llbracket |S| < k\rrbracket &(\min\{p^+(S,t), p^+(T,s)\} - \sum_{i\in [N]\backslash (S\cup T)}|p^+(S,i) - p^+(T,i)|)) = 1 - {1 - \alpha_2\over 2N}.
\end{align*}
Let $\alpha = \max_{(S,T)\in E}\{\alpha_1,\alpha_2\}$. If $\alpha < 1$, with Lemma~\ref{lem:pathcoupling} we have
\begin{equation*}
\tau(\eps) \le {2N\log(k/\eps)\over 1 - \alpha}.\qedhere
\end{equation*}

\end{proof}

\section{Proof of Thm.~\ref{thm:rayleigh}}\label{app:rayleigh}

We recall key aspects of strongly Rayleigh distributions, on which our proof of fast mixing depends\footnote{Part of the material is drawn from~\cite{borcea2009negative}, we include it for self-containness.}. Let $\pi$ be a probability distribution on $\{0,1\}^N$, its \emph{generating polynomial} is defined as
\begin{align*}
f_\pi (z) = \sum_{S\in [N]} \pi(S) z^S,
\end{align*}
where $z = (z_1,\ldots, z_N)$ and $z^S = \prod_{i\in S} z_i$. One of useful properties of such polynomial is their stability. A polynomial $f\in\mathbb{C}[z_1,\ldots, z_N]$ is called \emph{stable} if $f(z)\ne 0$ whenever $\ci\cm(z_j) > 0$ for $j\in [N]$. A stable polynomial with all real coefficients is called \emph{real stable}. 

Strongly Rayleigh distribution is defined upon properties of its generating polynomial: A distribution $\pi$ is called \emph{strongly Rayleigh} if its generating polynomial $f_\pi$ is (real) stable.

One of common manipulations on distributions over $\{0,1\}^N$ is symmetric homogenization, where one construct distributions on $\{0,1\}^{2N}$ such that their marginal distribution on $[N]$ is the same as the original ones on $\{0,1\}^N$.
\begin{definition}[Symmetric Homogenization]\label{def:sh}
Given $\pi$ on $\{0,1\}^N$, define a new distribution $\pi_{sh}$ on $\{0,1\}^{2N}$ called the \emph{symmetric homogenization} of $\pi$ by
\begin{align*}
\pi_{sh}(S) = \left\{\begin{array}{cc}
\pi(S\cap [N]){N \choose S\cap [N]}^{-1} & \text{if } |S| = N;\\
0 & \text{otherwise}.
\end{array}
\right.
\end{align*}
\end{definition}

The class of strongly Rayleigh distribution has been proved to be closed under symmetric homogenization: 

\begin{theorem}[Closure under Symmetric Homogenization~\cite{borcea2009negative}]\label{thm:closure}
If $\pi$ is strongly Rayleigh then so its symmetric homogenization $\pi_{sh}$.
\end{theorem}

Strongly Rayleigh distribution includes many distributions such as \dpp as special cases. Only recently, the Markov chain constructed for sampling from homogeneous strongly Rayleigh distribution has been proved to be rapidly mixing. 

\begin{theorem}[Rapid Mixing for Homogeneous Strongly Rayleigh~\cite{anari2016monte}]\label{thm:strongrayleigh}
For any strongly Rayleigh $k$-homogeneous probability distribution $\pi: \{0,1\}^{N}\to \mathbb{R}_+$, we have
\begin{align*}
\tau_{X_0}(\eps)\le 2k(2N-k)(\log\pi(X_0)^{-1} + \log\eps^{-1}).
\end{align*}
where $2N$ is the size of the ground set.
\end{theorem}

Now we are ready to prove our statement in Thm.~\ref{thm:rayleigh}.

\paragraph{Proof of Thm.~\ref{thm:rayleigh}}

Given a strongly Rayleigh distribution $\pi_\cc$, we construct its symmetric homogenization $\pi_{sh}$ as in Def.~\ref{def:sh}. By Thm.~\ref{thm:closure} we know that $\pi_{sh}$ is homogeneous strongly Rayleigh. Then it follows from Thm.~\ref{thm:strongrayleigh} that the base exchange Markov chain has its mixing time bounded as
\begin{align*}
(\tau_{sh})_{Y_0}(\eps)&\le 2 N^2 (\log (\pi_{sh}(Y_0))^{-1} + \log \eps^{-1})\\
&= 2 N^2 \left(\log {N\choose |X_0|} + \log (\pi_\cc(X_0))^{-1} + \log \eps^{-1}\right),
\end{align*}
where $Y_0\subseteq [2N]$, $|Y_0| = N$ and $X_0 = Y_0\cap V$.

We construct a base exchange Markov chain on $2N$ variables where we maintain a set $|R| = N$. In each iteration and with probability 0.5 we choose uniformly $s\in R$ and $t\in [2N]\backslash R$ and switch them with certain transition probabilities. Let $S = R\cap V$, $T = V\backslash R$, there are in total four possibilities for locations of $s$ and $t$:
\begin{enumerate}
\item With probability ${|S|(N-|S|)\over 2N^2}$, $s\in S$ and $t\in T$, and we switch assignment of $s$ and $t$ with probability $\min\{1,{\pi_{sh}(R\cup\{t\}\backslash\{s\})\over \pi_{sh}(R)}\} = \min\{1,{\pi_\cc(S\cup\{t\}\backslash\{s\})\over \pi_\cc(S)}\}$. This is equivalent to switching elements between $S$ and $T$;
\item With probability ${|S|(N-|S|)\over 2N^2}$, $s\notin S$ and $t\notin T$, and switch with probability $\min\{1,{\pi_\cc(S\cup\{t\})\over \pi_\cc(S)}\times {|S|+1\over N-|S|}\}$. This is equivalent to doing nothing to $S$;
\item With probability ${|S|^2\over 2N^2}$, $s\in S$ and $t\notin T$, and we switch with probability $\min\{1,{\pi_\cc(S\backslash\{s\})\over \pi_\cc(S)}\times {|S|\over N-|S|+1}\}$. This is equivalent to deleting elements from $S$;
\item With probability ${(N-|S|)^2\over 2N^2}$, $s\notin S$ and $t\in T$, and switch with probability $\min\{1,{\pi_\cc(S\cup\{t\})\over \pi_\cc(S)}\times {|S|+1\over N-|S|}\}$. This is equivalent to adding elements to $S$.
\end{enumerate}
Constructing the chain in the same manner but only maintaining $S = R\cap [N]$ will result in Algo.~\ref{algo:rayleigh}, while the mixing time stays unchanged.

\section{Supplementary Experiments}

\subsection{Varying $\delta$}\label{app:sec:delta}

\begin{figure}[h!]
\centering
	\begin{subfigure}{.32\textwidth}
	\centering
	\includegraphics[width=\textwidth]{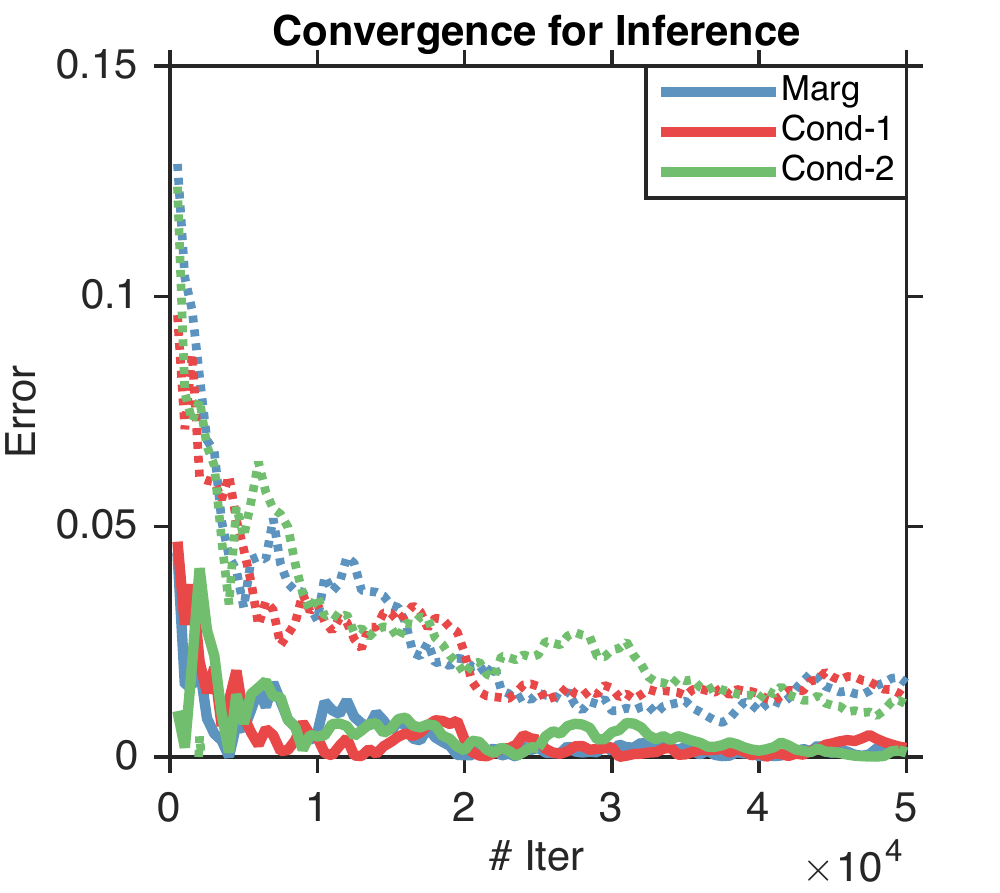}
	\caption{}
	\label{}
	\end{subfigure}%
	\begin{subfigure}{.32\textwidth}
	\centering
	\includegraphics[width=\textwidth]{ailerons_err_20_ising_fixpart_5_beta_3_05_05_50000}
	\caption{}
	\end{subfigure}%
	\begin{subfigure}{.32\textwidth}
	\centering
	\includegraphics[width=\textwidth]{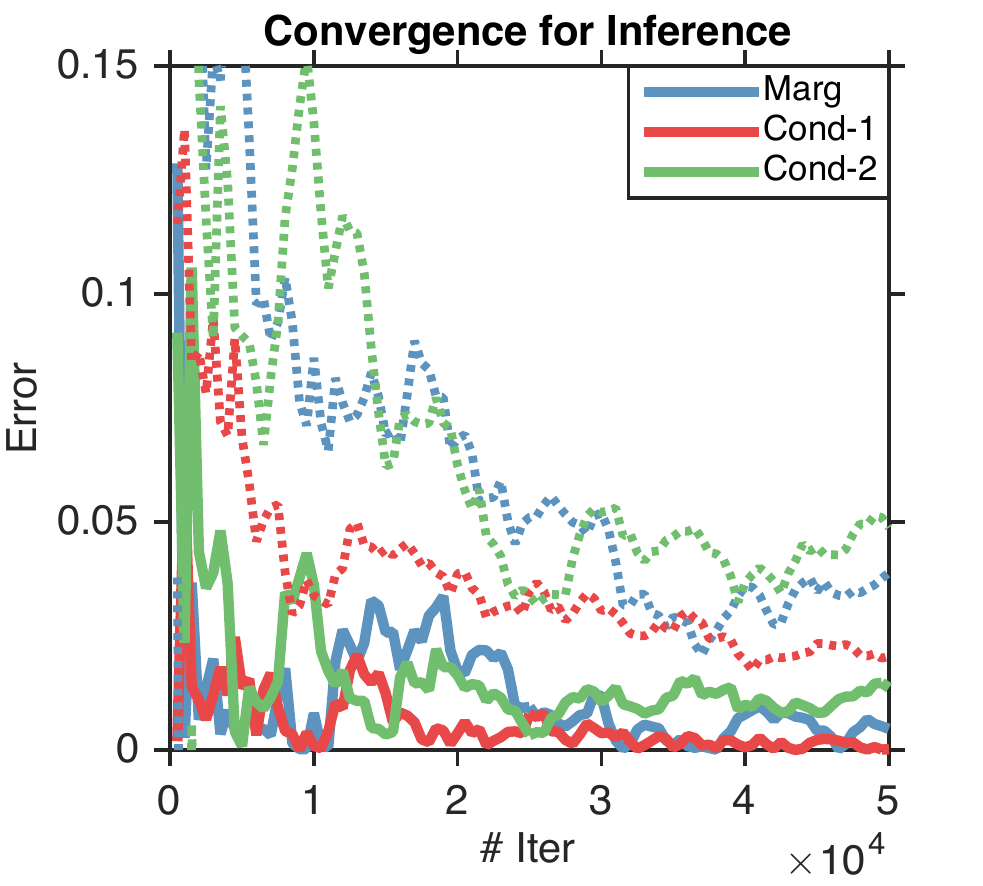}
	\caption{}
	\end{subfigure}
\caption{Convergence of marginal (\texttt{Marg}) and conditional~(\texttt{Cond-1} and \texttt{Cond-2}, conditioned on $1$ and $2$ other variables) probabilities of a single variable in a 20-variable Ising model. We fix $\beta=3$ and vary $\delta$ as (a) $\delta = 0.2$, (b) $\delta = 0.5$ and (c) $\delta = 0.8$. Full lines show the means and dotted lines the standard deviations of estimations.}
\label{fig:app:errdelta}
\end{figure}

\begin{figure}[h!]
\centering
	\begin{subfigure}{.32\textwidth}
	\centering
	\includegraphics[width=\textwidth]{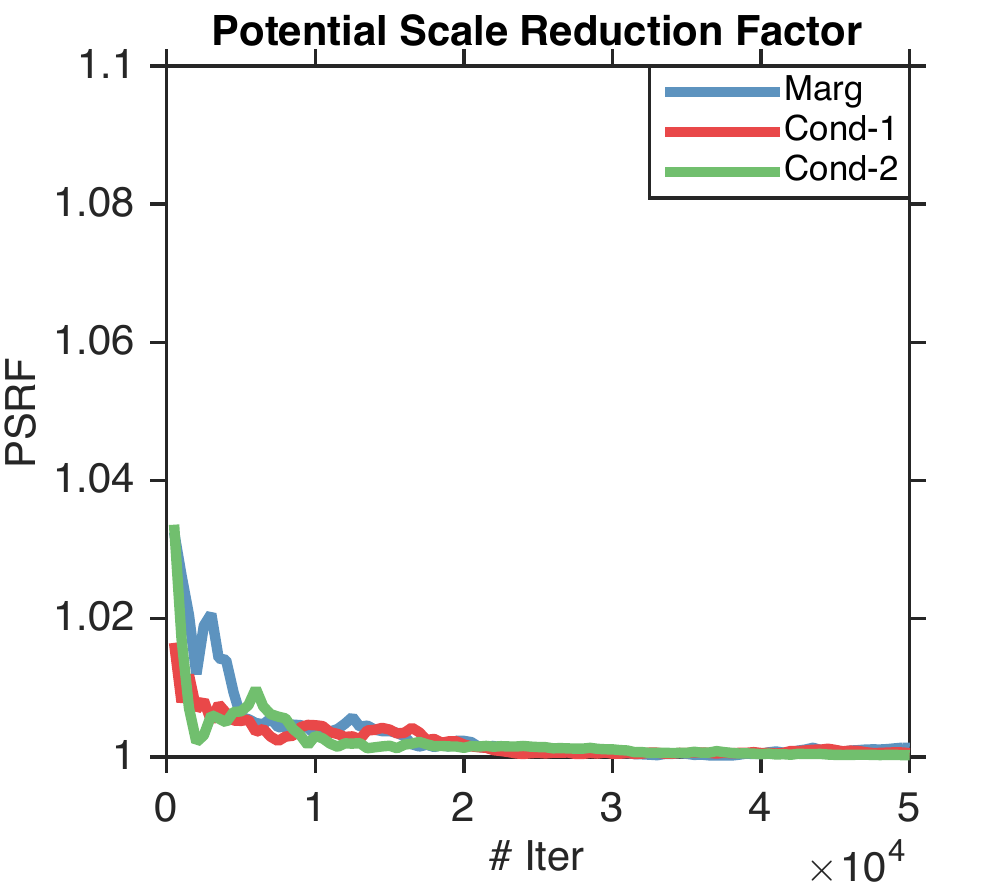}
	\caption{}
	\end{subfigure}%
	\begin{subfigure}{.32\textwidth}
	\centering
	\includegraphics[width=\textwidth]{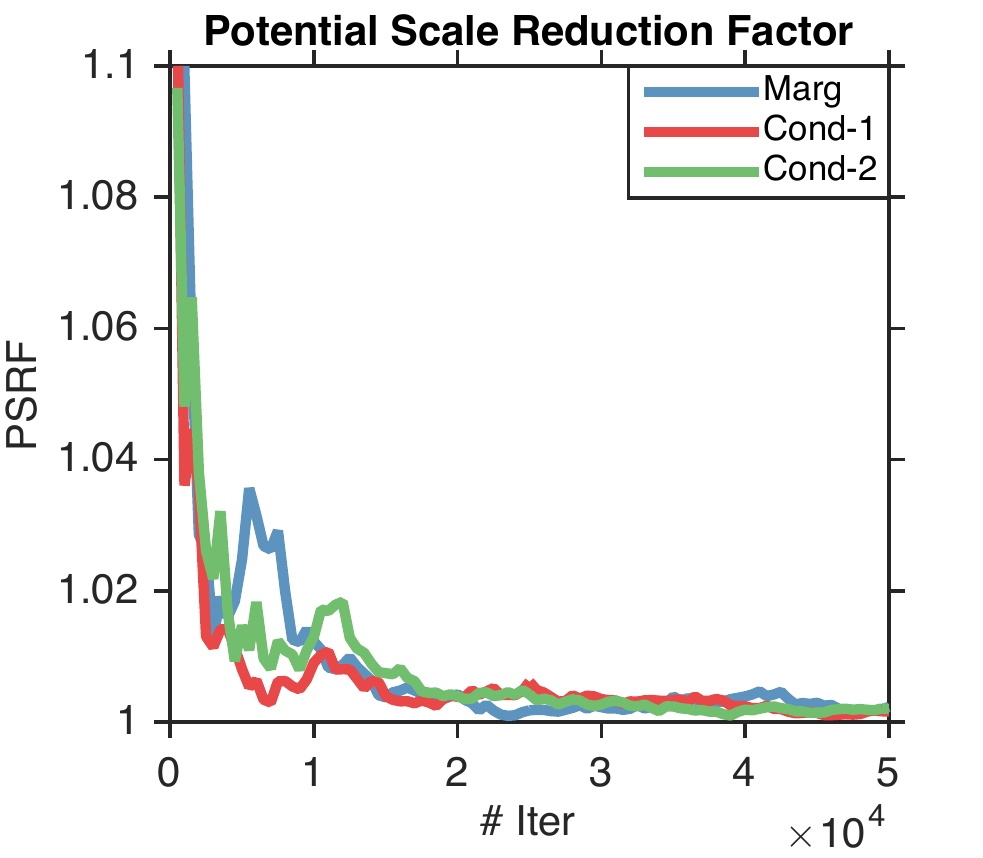}
	\caption{}
	\end{subfigure}%
	\begin{subfigure}{.32\textwidth}
	\centering
	\includegraphics[width=\textwidth]{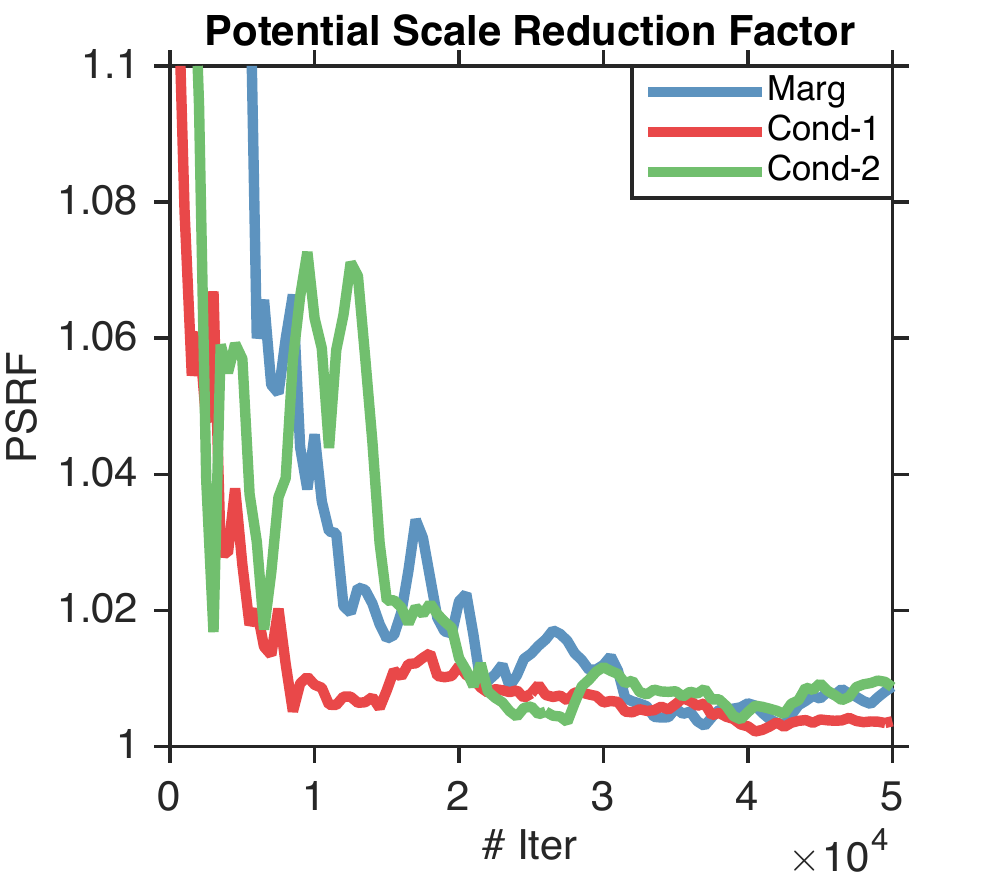}
	\caption{}
	\end{subfigure}
\caption{PSRF of each set of chains in Fig.~\ref{fig:app:errdelta} with $\beta=3$ and (a) $\delta = 0.2$; (b) $\delta = 0.5$ and (c) $\delta = 0.8$.}
\label{fig:app:convdelta}
\end{figure}

We run 20-variable chain-structured Ising model on partition matroid base of rank 5 with varying $\delta$'s. The results are shown in Fig.~\ref{fig:app:errdelta} and Fig.~\ref{fig:app:convdelta}. We observe that the approximate mixing time grows with $\delta$. 

\begin{figure}[h!]
\centering
	\begin{subfigure}{.32\textwidth}
	\centering
	\includegraphics[width=\textwidth]{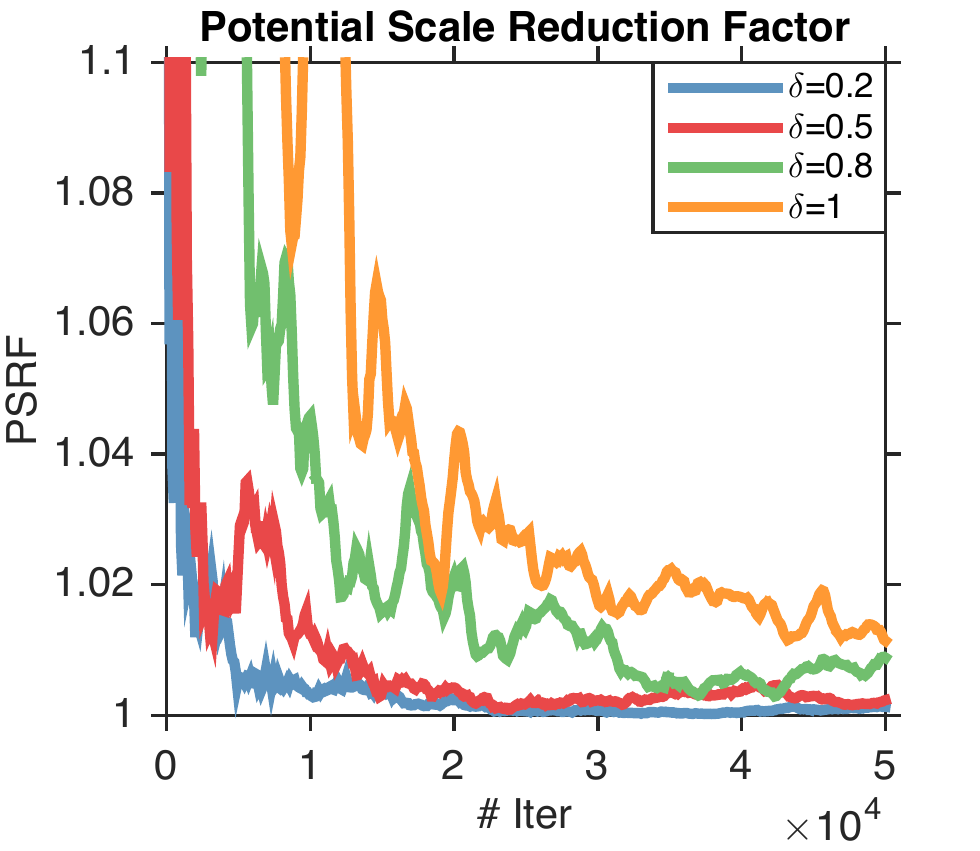}
	\caption{}
	\end{subfigure}%
	\begin{subfigure}{.32\textwidth}
	\centering
	\includegraphics[width=\textwidth]{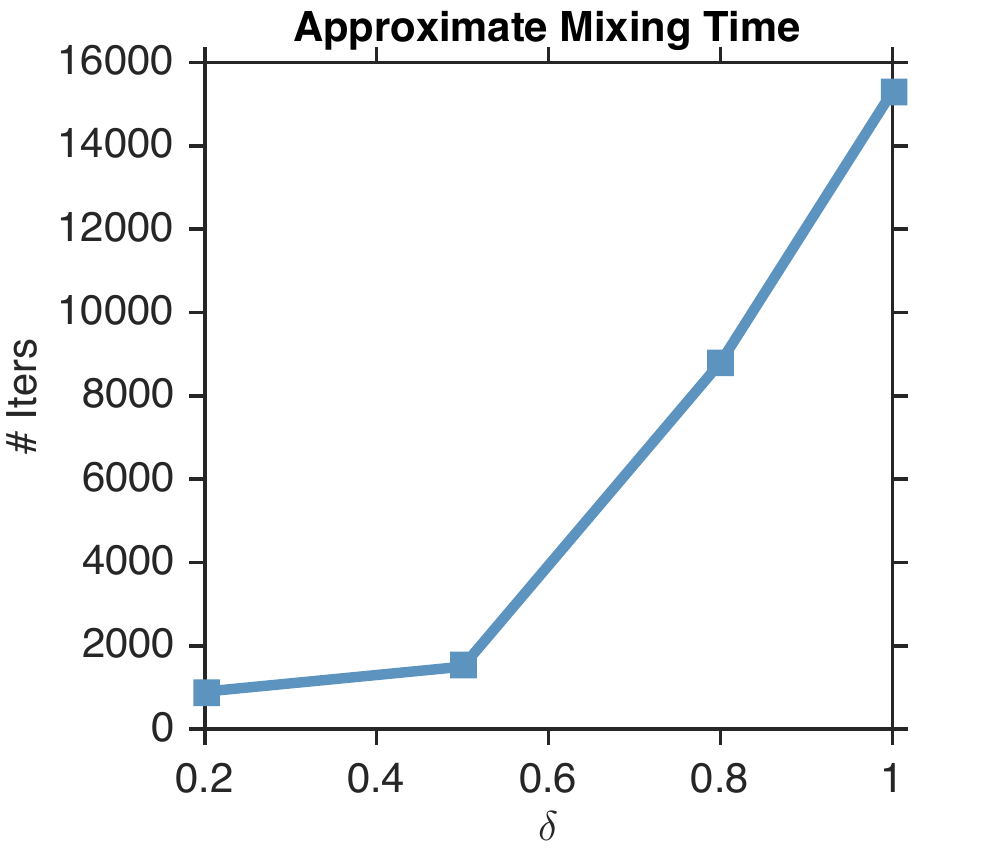}
	\caption{}
	\end{subfigure}
\caption{Comparisons of PSRF's for marginal estimations with different $\delta$'s. (a) PSRF's with different $\delta$'s and (b) the approximate mixing time estimated by thresholding PSRF at 1.05.}
\label{fig:large}
\end{figure}

\subsection{Varying $\beta$}\label{app:sec:beta}

\begin{figure}[h!]
\centering
	\begin{subfigure}{.32\textwidth}
	\centering
	\includegraphics[width=\textwidth]{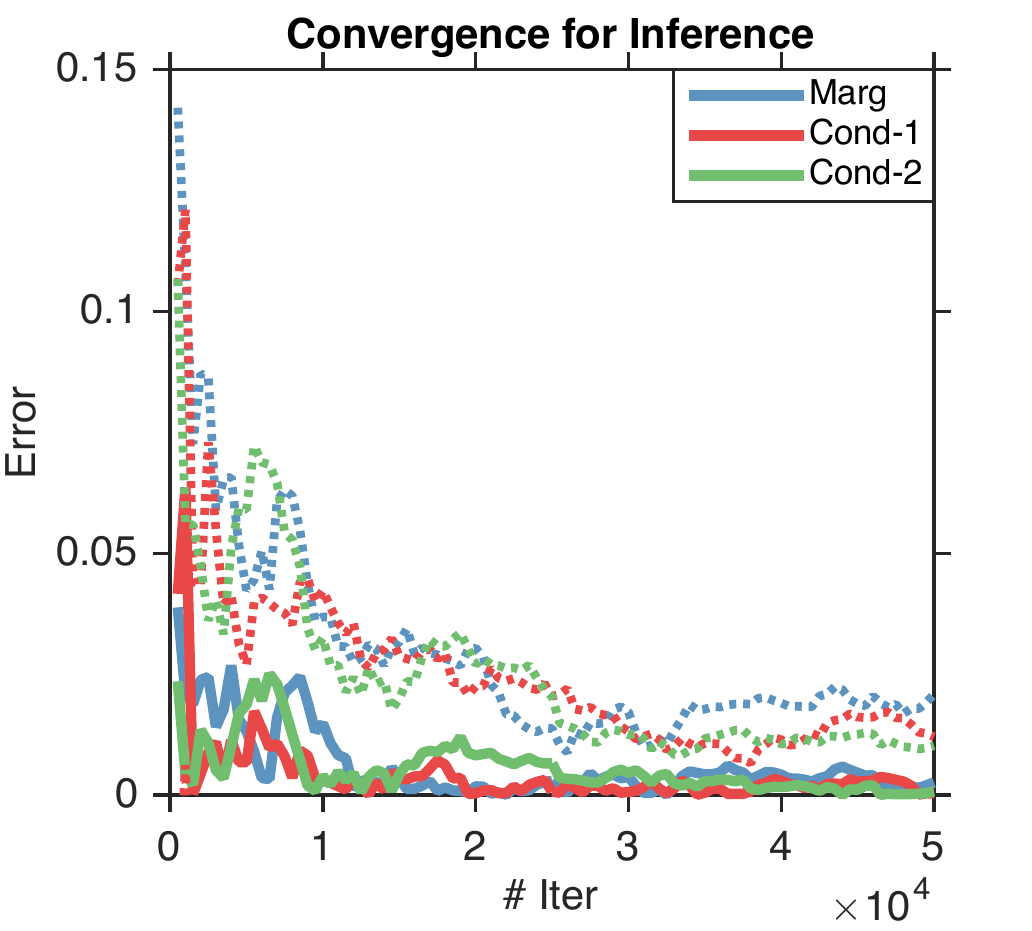}
	\caption{}
	\label{}
	\end{subfigure}%
	\begin{subfigure}{.32\textwidth}
	\centering
	\includegraphics[width=\textwidth]{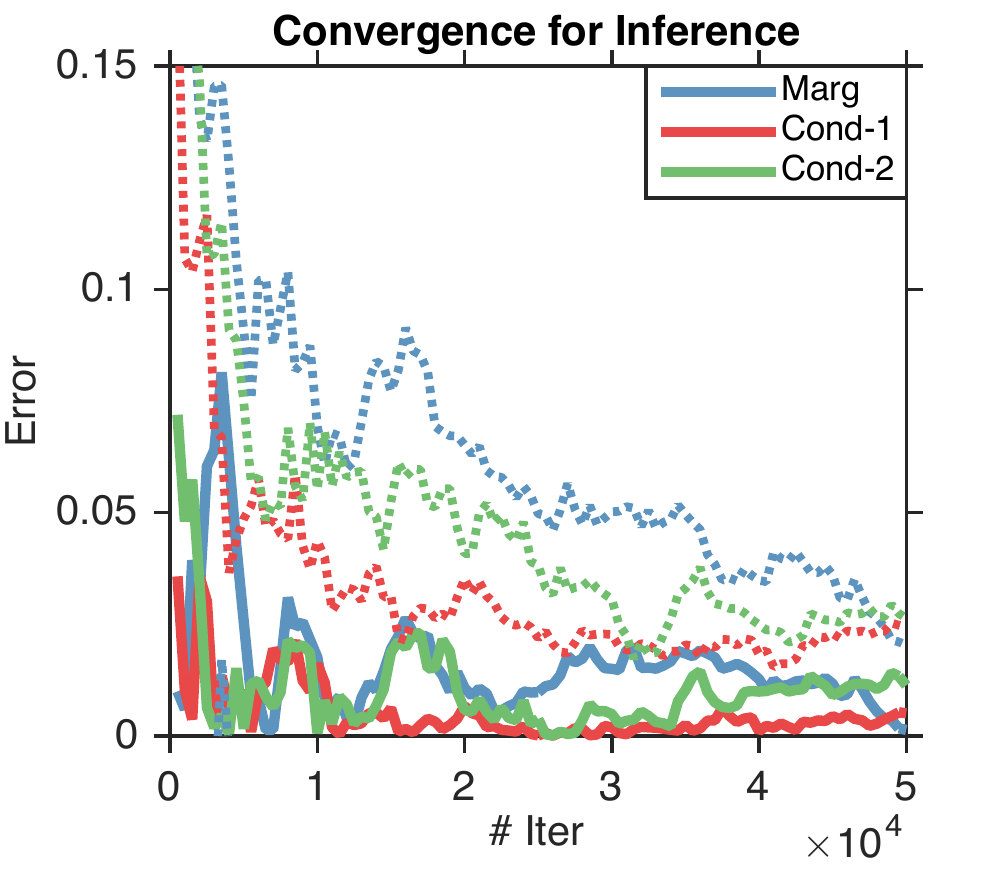}
	\caption{}
	\end{subfigure}%
	\begin{subfigure}{.32\textwidth}
	\centering
	\includegraphics[width=\textwidth]{ailerons_err_20_ising_fixpart_5_beta_3_50000}
	\caption{}
	\end{subfigure}
\caption{Convergence of marginal (\texttt{Marg}) and conditional~(\texttt{Cond-1} and \texttt{Cond-2}, conditioned on $1$ and $2$ other variables) probabilities of a single variable in a 20-variable Ising model. We fix $\delta=1$ and vary $\beta$ as (a) $\beta = 0.5$; (b) $\beta = 2$ and (c) $\beta = 3$. Full lines show the means and dotted lines the standard deviations of estimations.}
\label{fig:app:errbeta}
\end{figure}

\begin{figure}[h!]
\centering
	\begin{subfigure}{.32\textwidth}
	\centering
	\includegraphics[width=\textwidth]{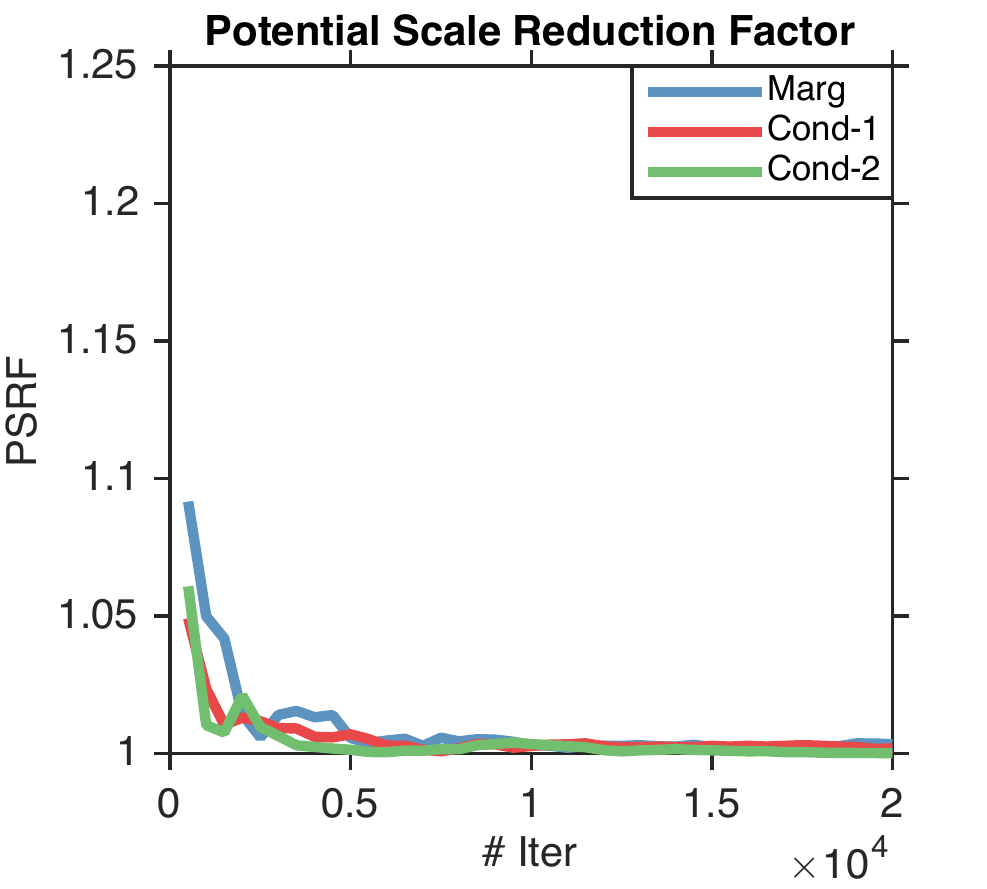}
	\caption{}
	\end{subfigure}%
	\begin{subfigure}{.32\textwidth}
	\centering
	\includegraphics[width=\textwidth]{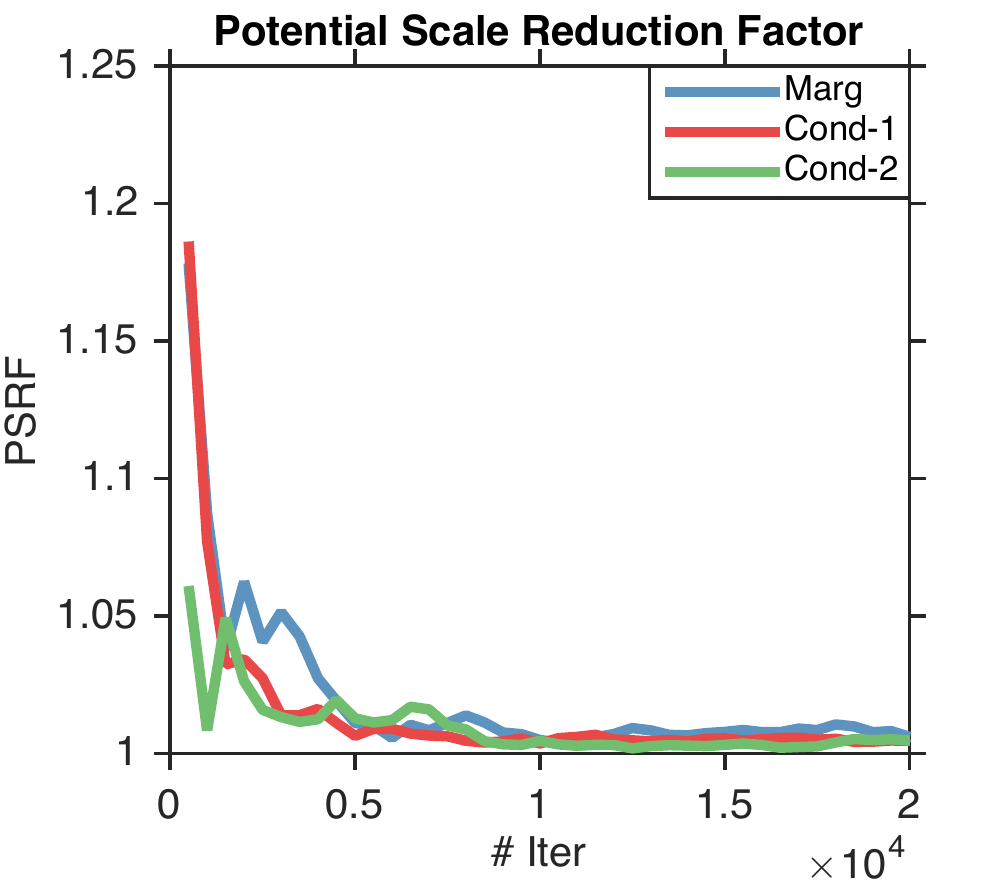}
	\caption{}
	\end{subfigure}%
	\begin{subfigure}{.32\textwidth}
	\centering
	\includegraphics[width=\textwidth]{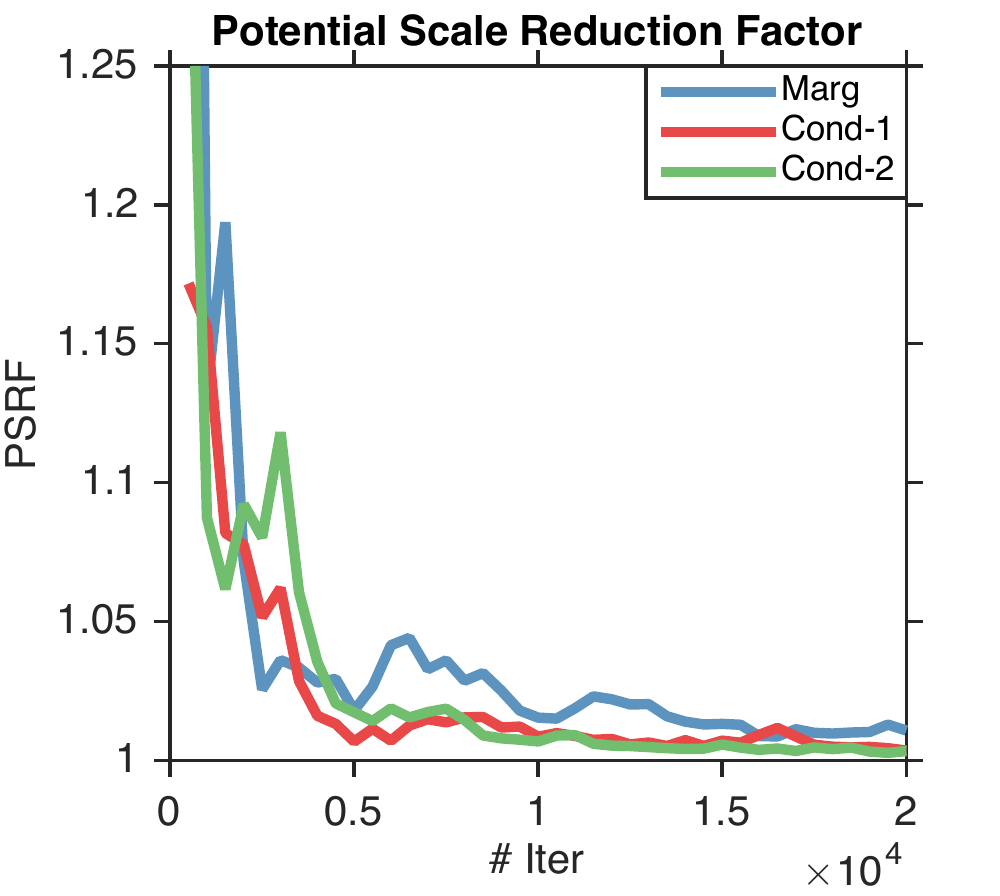}
	\caption{}
	\end{subfigure}
\caption{PSRF of each set of chains in Fig.~\ref{fig:app:errbeta} with $\delta=1$ and (a) $\beta = 0.5$; (b) $\beta = 2$ and (c) $\beta = 3$.}
\label{fig:app:convbeta}
\end{figure}

We run 20-variable chain-structured Ising model on partition matroid base of rank 5 with varying $\beta$'s. The results are shown in Fig.~\ref{fig:app:errbeta} and Fig.~\ref{fig:app:convbeta}. We observe that the approximate mixing time grows with $\beta$. 

\begin{figure}[h!]
\centering
	\begin{subfigure}{.32\textwidth}
	\centering
	\includegraphics[width=\textwidth]{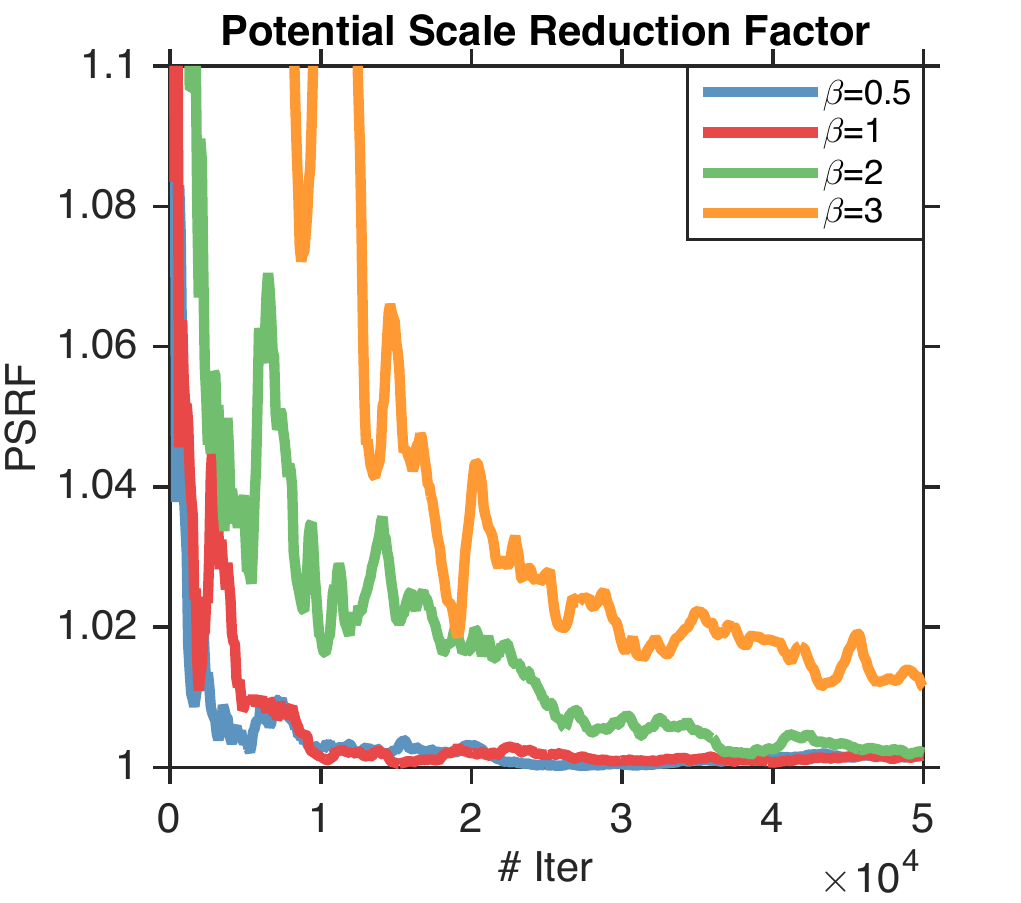}
	\caption{}
	\end{subfigure}%
	\begin{subfigure}{.32\textwidth}
	\centering
	\includegraphics[width=\textwidth]{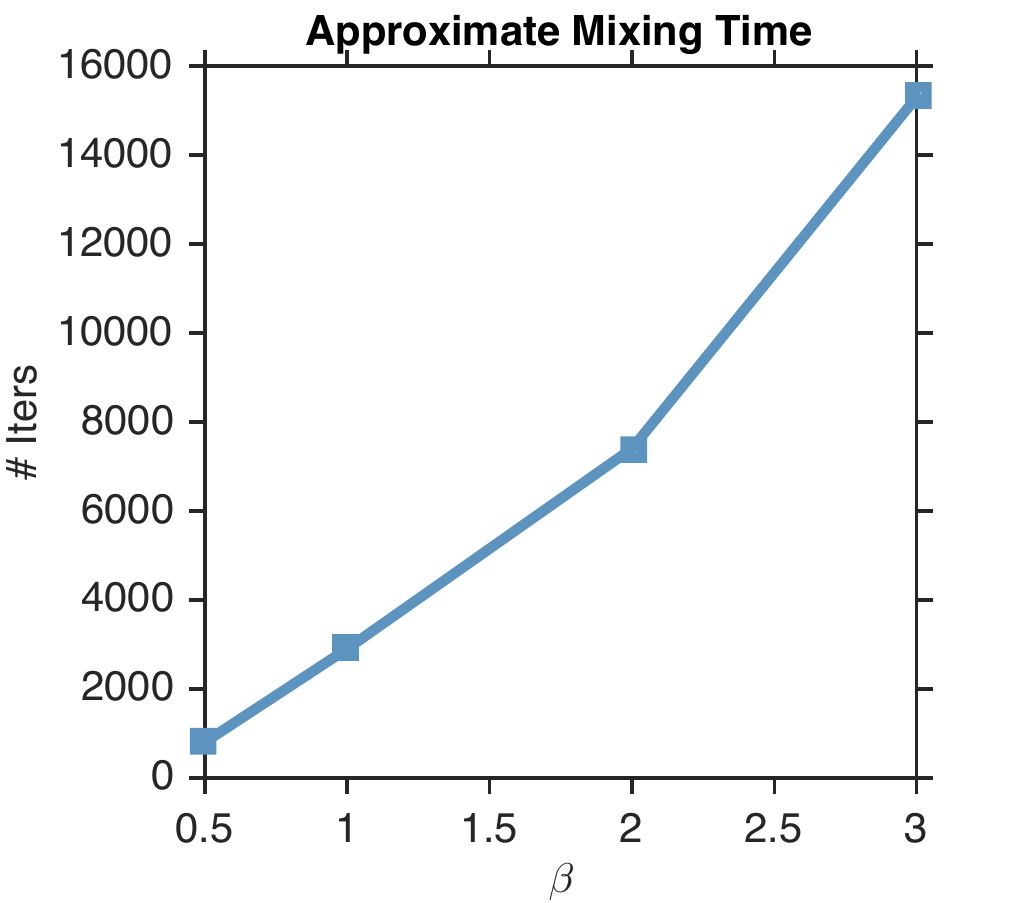}
	\caption{}
	\end{subfigure}
\caption{Comparisons of PSRF's for marginal estimations with different $\beta$'s. (a) PSRF's with different $\beta$'s and (b) the approximate mixing time estimated by thresholding of 1.05 on PSRF's.}
\label{fig:large}
\end{figure}

\subsection{Varying Data Sizes}\label{app:sec:size}

We run ($k$-)\dpp that is constrained to sample subsets from 1) partition matroid base and 2) uniform matroid with different data sizes $N$. 

\subsubsection{Partition Matroid Constraint}

\begin{figure}[h!]
\centering
	\begin{subfigure}{.32\textwidth}
	\centering
	\includegraphics[width=\textwidth]{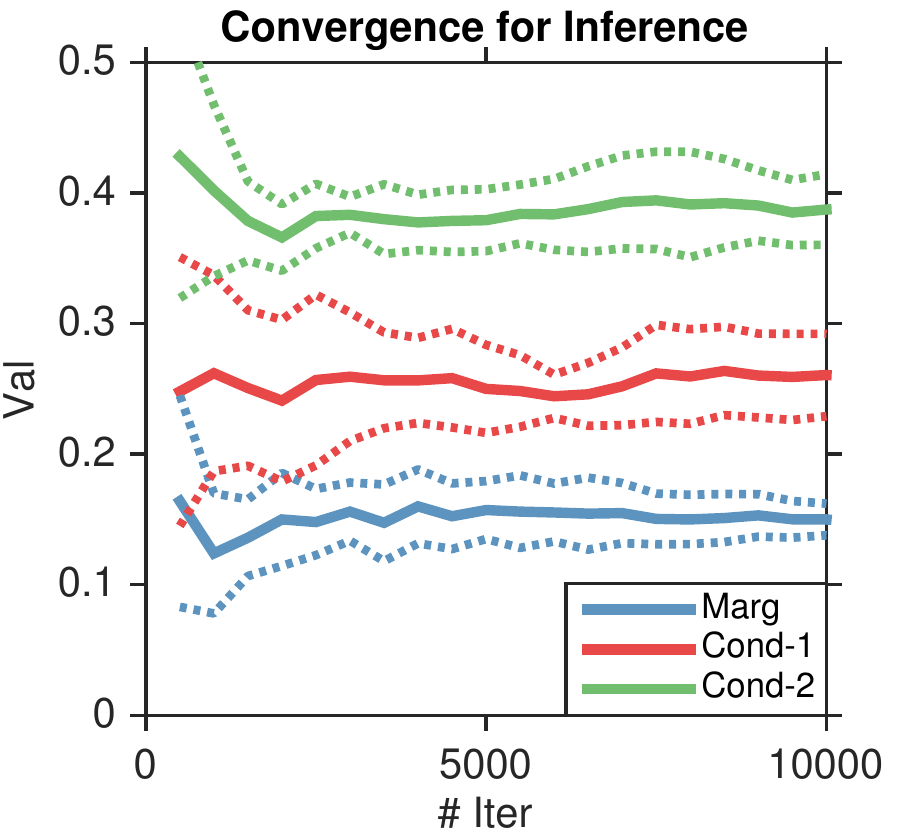}
	\caption{}
	\end{subfigure}%
	\begin{subfigure}{.32\textwidth}
	\centering
	\includegraphics[width=\textwidth]{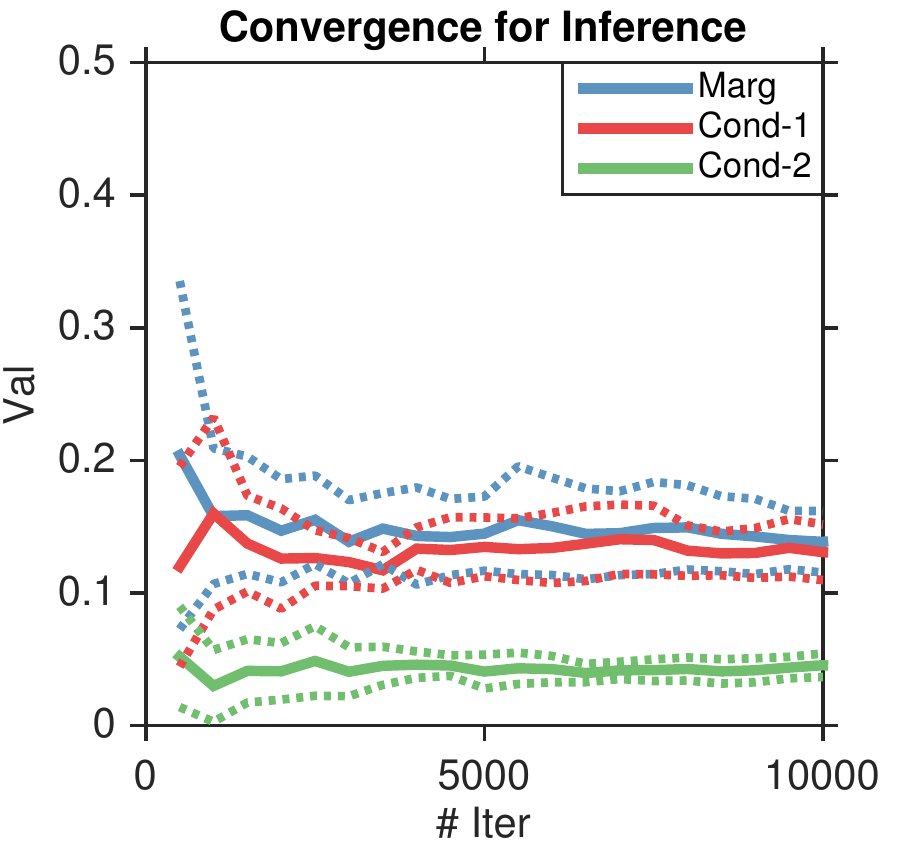}
	\caption{}
	\end{subfigure}%
	\begin{subfigure}{.32\textwidth}
	\centering
	\includegraphics[width=\textwidth]{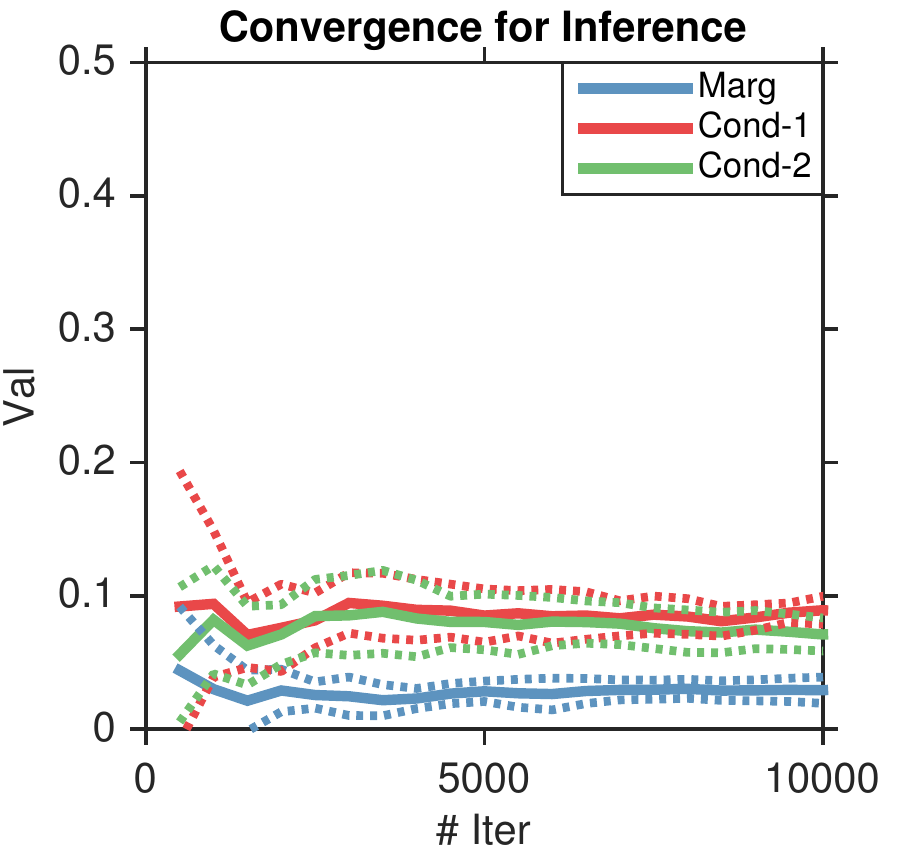}
	\caption{}
	\end{subfigure}
\caption{Convergence of marginal (\texttt{Marg}) and conditional (\texttt{Cond-1} and \texttt{Cond-2}, conditioned on $1$ and $2$ other variables) probabilities of a single variable in a $k$-\dpp on partition matroid base of rank 5, with (a) $N=20$; (b) $N=50$ and (c) $N=100$. Full lines show the means and dotted lines the standard deviations of estimations.}
\label{fig:app:partn}
\end{figure}

\begin{figure}[h!]
\centering
	\begin{subfigure}{.32\textwidth}
	\centering
	\includegraphics[width=\textwidth]{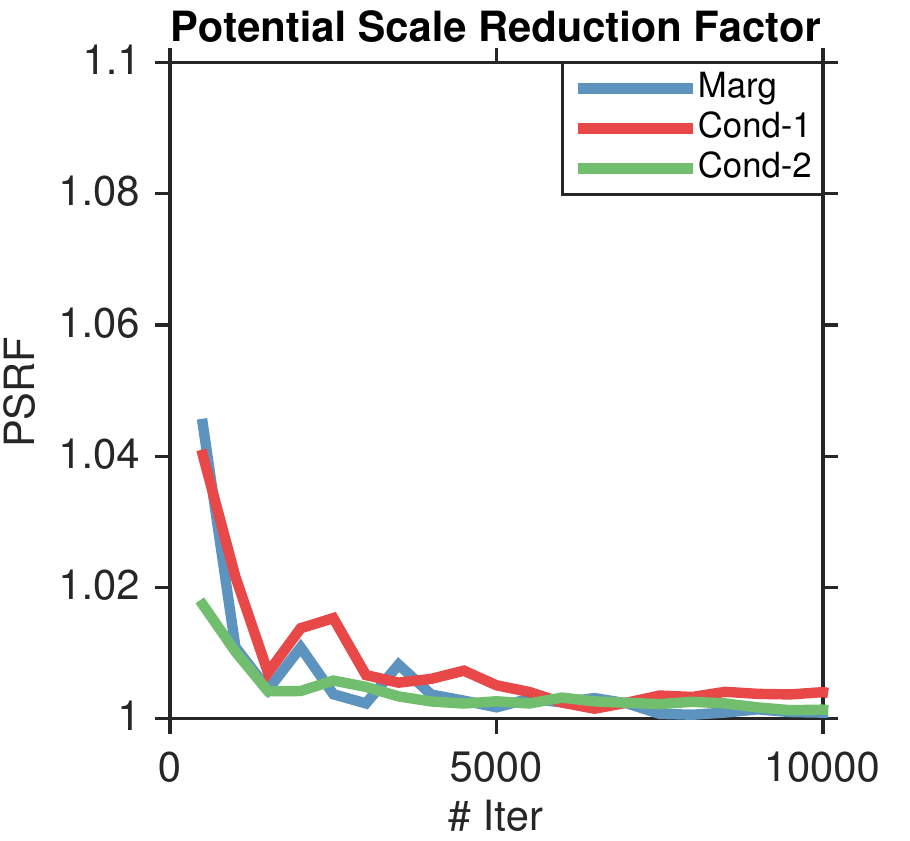}
	\caption{}
	\end{subfigure}%
	\begin{subfigure}{.32\textwidth}
	\centering
	\includegraphics[width=\textwidth]{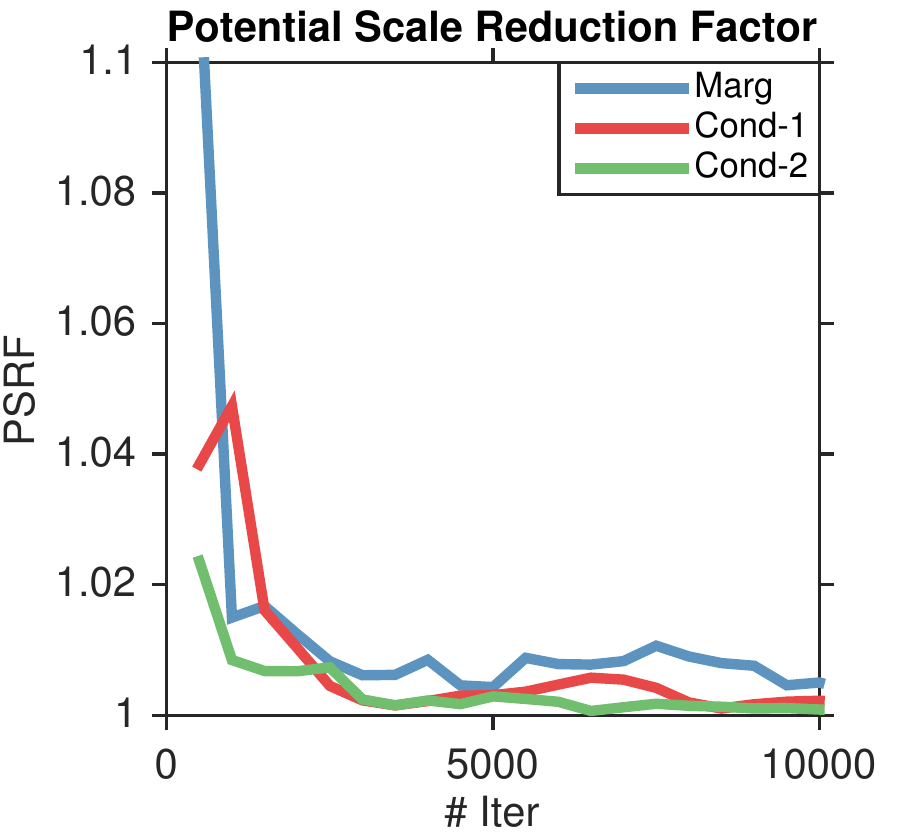}
	\caption{}
	\end{subfigure}%
	\begin{subfigure}{.32\textwidth}
	\centering
	\includegraphics[width=\textwidth]{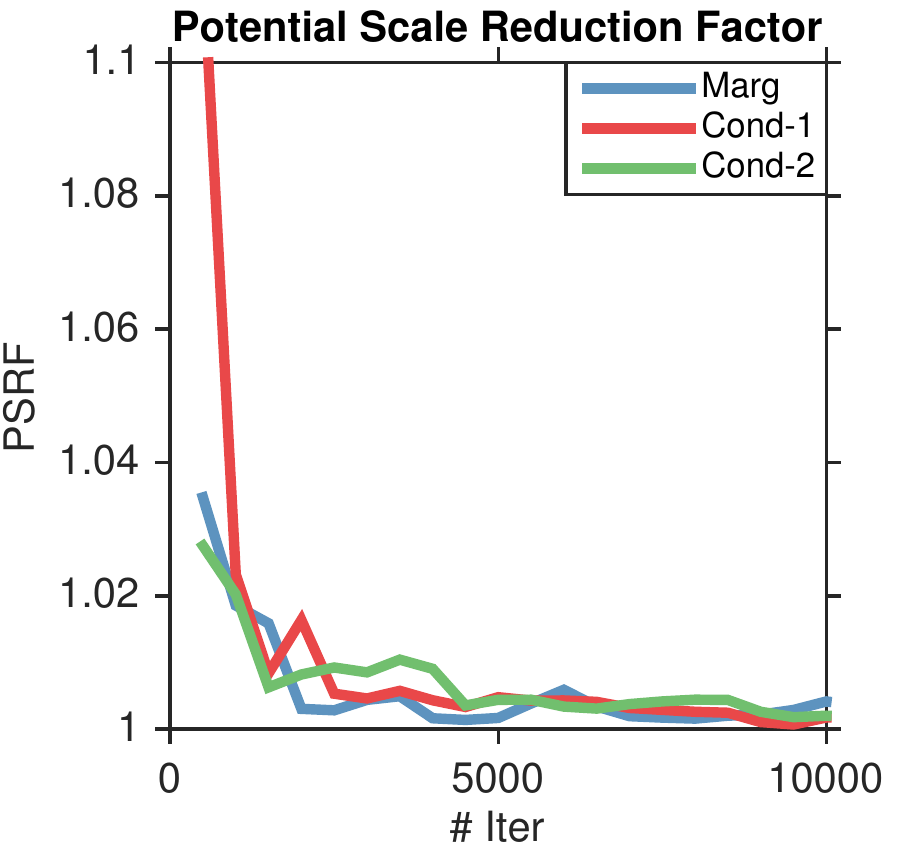}
	\caption{}
	\end{subfigure}
\caption{PSRF of marginal (\texttt{Marg}) and conditional (\texttt{Cond-1} and \texttt{Cond-2}, conditioned on $5$ and $10$ other variables) probabilities of a single variable in a $k$-\dpp on partition matroid base of rank 5, with (a) $N=20$; (b) $N=50$ and (c) $N=100$.}
\label{fig:app:psrfn}
\end{figure}

The estimations for marginal and conditional distributions are shown in Fig.~\ref{fig:app:partn} and corresponding PSRF's are shown in Fig.~\ref{fig:app:psrfn}. We observe that the estimation becomes stable faster when $N$ is small.

\subsubsection{Uniform Matroid Constraint}

\begin{figure}[h!]
\centering
	\begin{subfigure}{.32\textwidth}
	\centering
	\includegraphics[width=\textwidth]{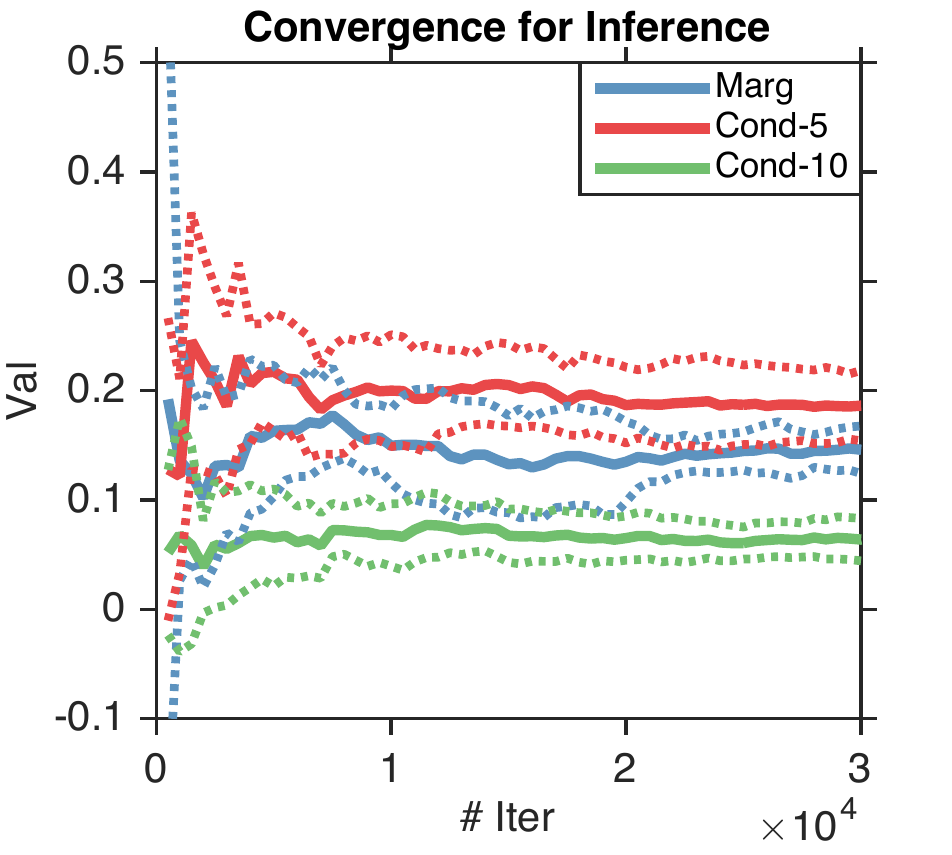}
	\caption{}
	\end{subfigure}%
	\begin{subfigure}{.32\textwidth}
	\centering
	\includegraphics[width=\textwidth]{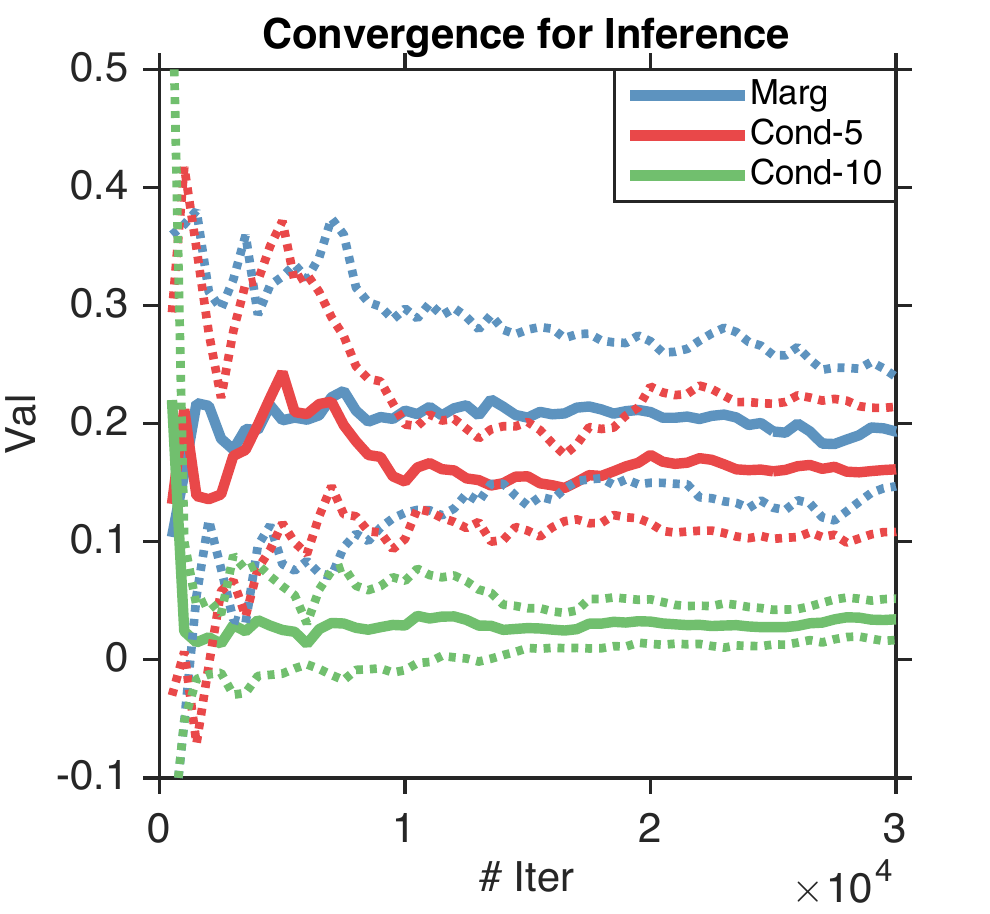}
	\caption{}
	\end{subfigure}%
	\begin{subfigure}{.32\textwidth}
	\centering
	\includegraphics[width=\textwidth]{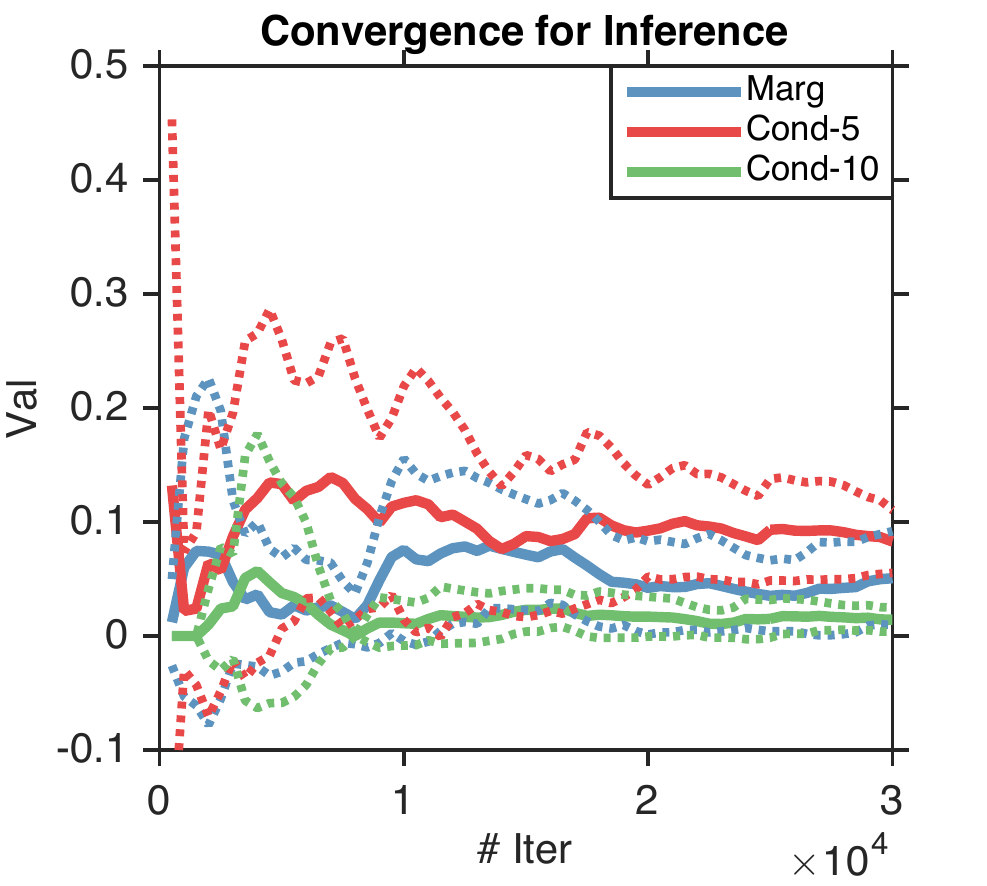}
	\caption{}
	\end{subfigure}
\caption{Convergence of marginal (\texttt{Marg}) and conditional (\texttt{Cond-5} and \texttt{Cond-10}, conditioned on $5$ and $10$ other variables) probabilities of a single variable in a \dpp on uniform matroid of rank 30, with (a) $N=50$; (b) $N=100$ and (c) $N=200$. Full lines show the means and dotted lines the standard deviations of estimations.}
\label{fig:app:errn}
\end{figure}

\begin{figure}[h!]
\centering
	\begin{subfigure}{.32\textwidth}
	\centering
	\includegraphics[width=\textwidth]{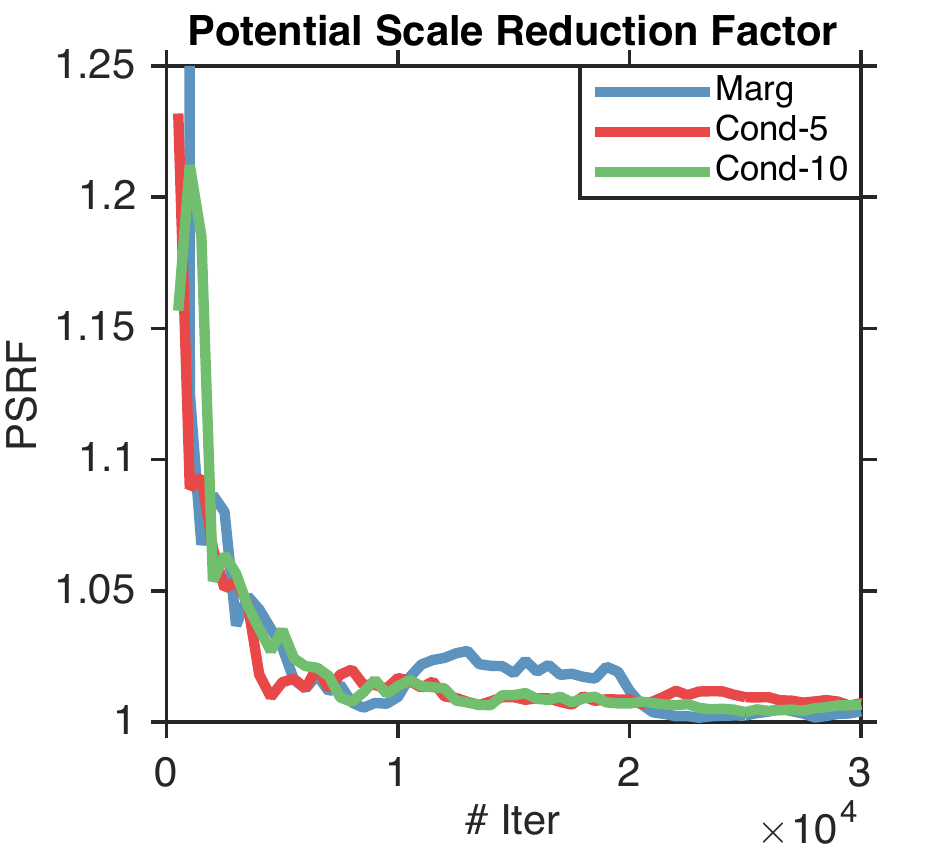}
	\caption{}
	\end{subfigure}%
	\begin{subfigure}{.32\textwidth}
	\centering
	\includegraphics[width=\textwidth]{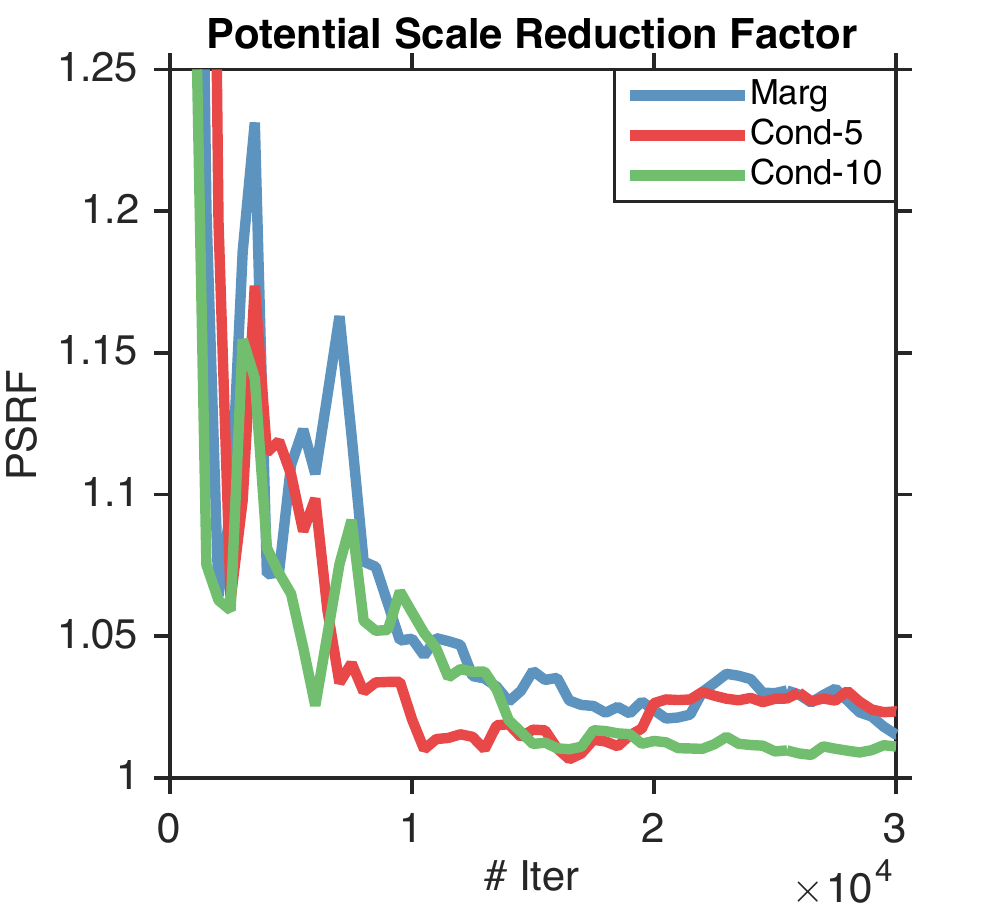}
	\caption{}
	\end{subfigure}%
	\begin{subfigure}{.32\textwidth}
	\centering
	\includegraphics[width=\textwidth]{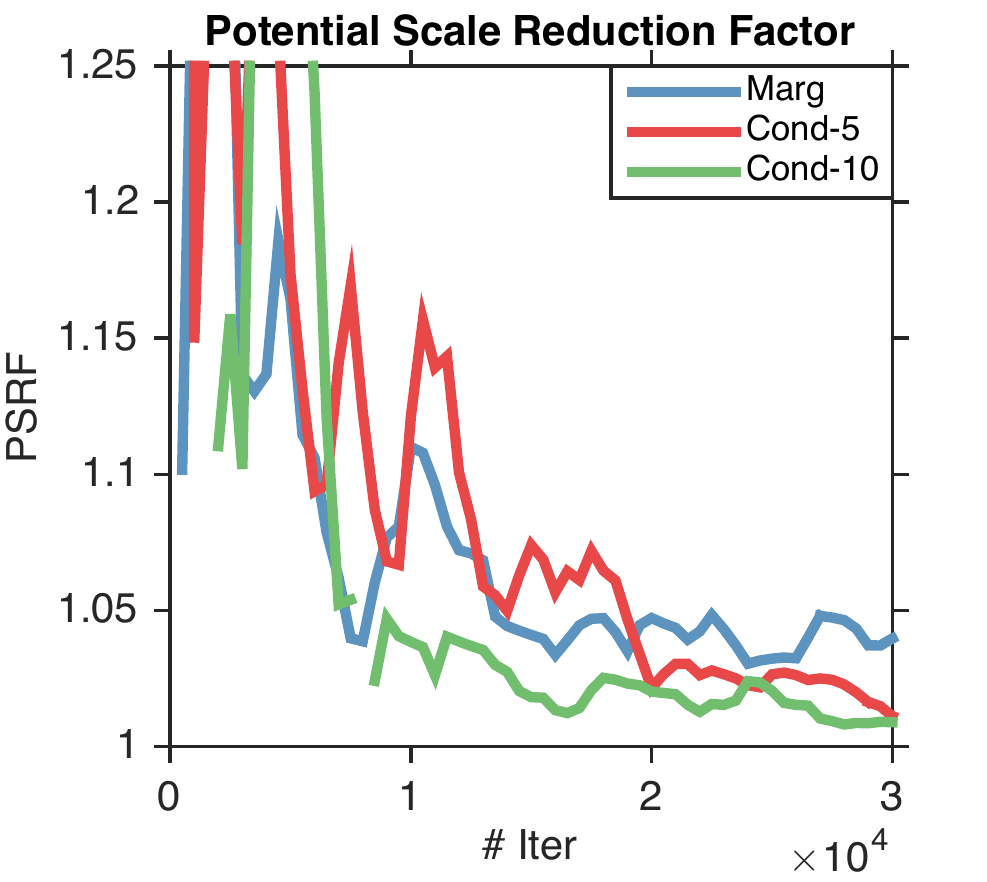}
	\caption{}
	\end{subfigure}
\caption{PSRF of marginal (\texttt{Marg}) and conditional (\texttt{Cond-5} and \texttt{Cond-10}, conditioned on $5$ and $10$ other variables) probabilities of a single variable in a \dpp on uniform matroid of rank 30, with (a) $N=50$; (b) $N=100$ and (c) $N=200$.}
\label{fig:app:convn}
\end{figure}

The estimations for marginal and conditional distributions are shown in Fig.~\ref{fig:app:errn} and corresponding PSRF's are shown in Fig.~\ref{fig:app:convn}. We observe the same thing as mentioned before, that the estimation becomes stable faster when $N$ is small.

\end{appendix}

\end{document}